\pdfoutput=1
\documentclass[11pt]{article}
\usepackage[ruled,linesnumbered,noend,boxed,noline]{algorithm2e}%用于显示伪代码
\usepackage{amsmath}   % 基础数学扩展
\usepackage{amssymb}   % 数学符号（如\mathbb{I}）
\usepackage{mathtools}

\usepackage[table]{xcolor}%for \rowcolor[HTML]{faf0e6}
\usepackage[utf8]{inputenc} % allow utf-8 input
\usepackage[T1]{fontenc}    % use 8-bit T1 fonts
\usepackage{hyperref}       % hyperlinks
\usepackage{url}            % simple URL typesetting
\usepackage{booktabs}       % professional-quality tables
\usepackage{amsfonts}       % blackboard math symbols
\usepackage{nicefrac}       % compact symbols for 1/2, etc.
\usepackage{microtype}      % microtypography
\usepackage{xcolor}         % colors
\usepackage[normalem]{ulem}
\useunder{\uline}{\ul}{}
\usepackage{graphicx}
\usepackage{multirow}
\usepackage{longtable}
\usepackage{wrapfig}
\usepackage{subcaption}
\usepackage{caption}
\usepackage{CJKutf8}
\usepackage{float} % allow add graph at specific places
\usepackage{tikz}

\usepackage{cleveref}%for \Cref{fig:GUI-explorer}~(c)

\usepackage{tcolorbox} % for visualise prompt
\definecolor{beigebg}{RGB}{253,249,238} % Light beige
\definecolor{argcolor}{RGB}{70,130,180} % Steel blue for arg-inputs
\newcommand{\arginputs}[1]{\textcolor{argcolor}{#1}}

\tcbset{
  mypromptstyle/.style={
    colback=beigebg,
    colframe=beigebg,
    boxrule=0pt,
    sharp corners,
    fontupper=\ttfamily\scriptsize,
    before=\vspace{0.5em},
    after=\vspace{0.5em},
  }
}
\usepackage{xparse}
\NewDocumentEnvironment{prompt}{O{}}
{
  \begin{tcolorbox}[mypromptstyle]
  \obeylines  % 让换行符（回车）生效
  \obeyspaces % 让多个空格生效
  \parindent=0pt  % 取消自动缩进
}
{
  \end{tcolorbox}
}

\newcommand{\secref}[1]{Section~\ref{#1}} % Simple macro for section reference
\newcommand{\appref}[1]{Appendix~\ref{#1}}

\DeclareMathOperator*{\argmax}{arg\,max}

\usepackage{acl}%final版本

\usepackage{times}
\usepackage{latexsym}

\usepackage[T1]{fontenc}
\usepackage[utf8]{inputenc}

\usepackage{microtype}

\usepackage{inconsolata}

\usepackage{graphicx}

\title{GUI-explorer: Autonomous Exploration and Mining of Transition-aware Knowledge for GUI Agent}

\author{
  \textbf{Bin Xie\textsuperscript{1}},
  \textbf{Rui Shao\textsuperscript{1\textdagger}},
  \textbf{Gongwei Chen\textsuperscript{1\textdagger}},
  \textbf{Kaiwen Zhou\textsuperscript{2}},
\\
  \textbf{Yinchuan Li\textsuperscript{2}},
  \textbf{Jie Liu\textsuperscript{1}},
  \textbf{Min Zhang\textsuperscript{1}},
  \textbf{Liqiang Nie \textsuperscript{1}}%,
\\
\\
  \textsuperscript{1}Harbin Institute of Technology, Shenzhen,
  \textsuperscript{2}Huawei Noah’s Ark Lab
}

\begin{document}
\maketitle

\begin{abstract}
GUI automation faces critical challenges in dynamic environments. MLLMs suffer from two key issues: misinterpreting UI components and outdated knowledge. Traditional fine-tuning methods are costly for app-specific knowledge updates.
We propose GUI-explorer, a training-free GUI agent that incorporates two fundamental mechanisms: 
\textbf{(1) Autonomous Exploration of Function-aware Trajectory}. To comprehensively cover all application functionalities, we design a \textbf{Function-aware Task Goal Generator} that automatically constructs exploration goals by analyzing GUI structural information (e.g., screenshots and activity hierarchies). This enables systematic exploration to collect diverse trajectories.
\textbf{(2) Unsupervised Mining of Transition-aware Knowledge}. To establish precise screen-operation logic, we develop a \textbf{Transition-aware Knowledge Extractor} that extracts effective screen-operation logic through unsupervised analysis the state transition of structured interaction triples (observation, action, outcome). This eliminates the need for human involvement in knowledge extraction.
With a task success rate of 53.7\% on SPA-Bench and 47.4\% on AndroidWorld, GUI-explorer shows significant improvements over SOTA agents.
It requires no parameter updates for new apps.
GUI-explorer is open-sourced and publicly available at \url{https://github.com/JiuTian-VL/GUI-explorer}.
\end{abstract}

{\let\thefootnote\relax\footnotetext{\textsuperscript{\textdagger} Corresponding authors. shaorui@hit.edu.cn, chengongwei@hit.edu.cn.}}

\section{Introduction}

Automation in graphical user interfaces (GUIs) has rapidly advanced~\citep{language-agent-tutorial}. This progress is driven by foundational models like large language models (LLMs)~\citep{touvron2023llama,achiam2023gpt,qwen2.5} and multimodal large language models (MLLMs)~\citep{hurst2024gpt,Chen_2024_CVPR,shao2024detecting,gemini_2,shao2023detecting,wei2024lion,shen2024mome}. These innovations enable agents~\citep{zheng2024seeact,10.1145/3706598.3713600,NEURIPS2024_0520537b,li2025optimus,ye2024cat,li2025optimus2} to handle tasks. They require no extensive fine-tuning or pretraining. This demonstrates their potential for diverse applications.
\begin{figure}[t]
  \includegraphics[width=\linewidth]{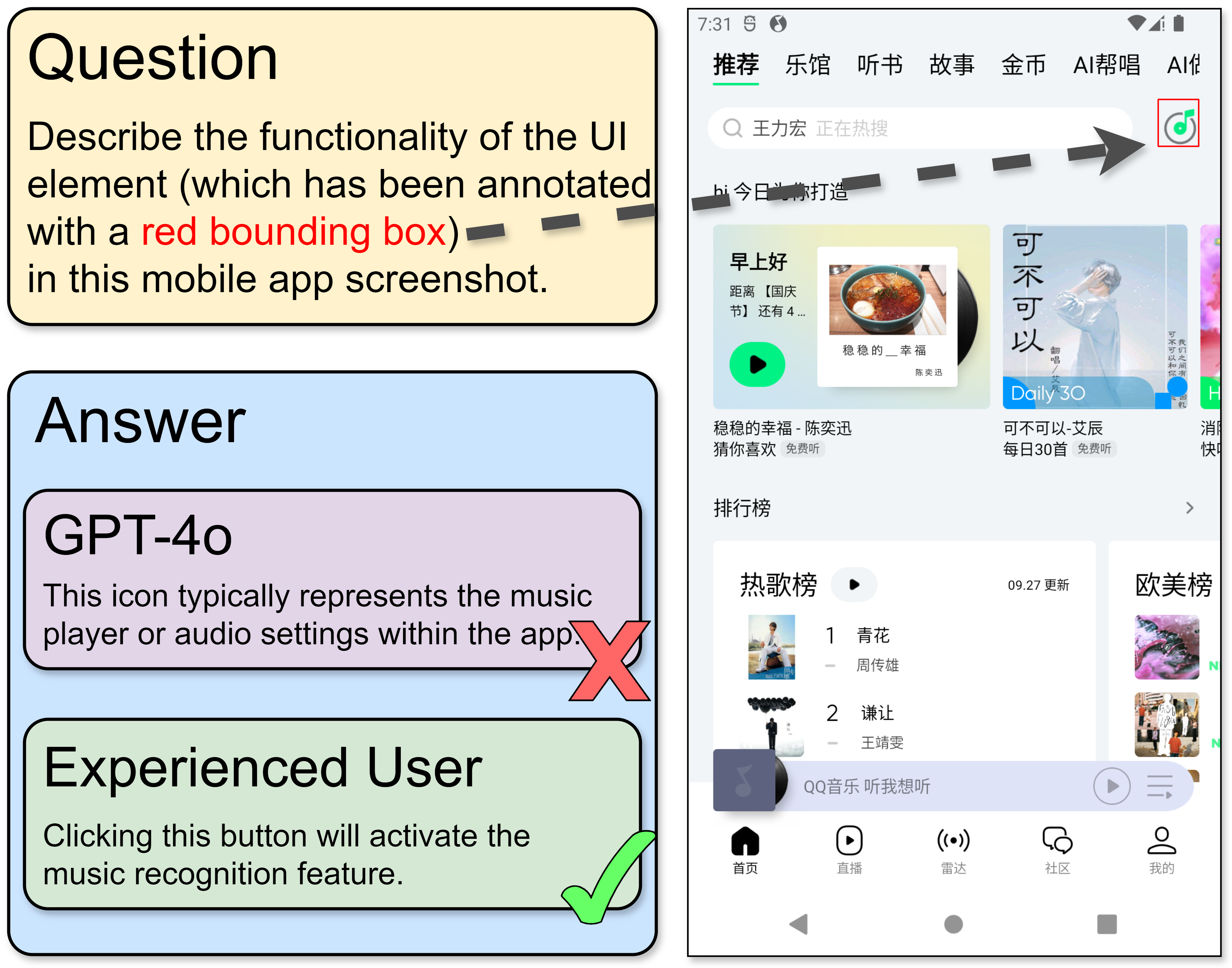}
  \caption{Comparison of GPT-4o and an user's interpretation of a UI element in QQ Music\footnotemark. The red-bounded icon in the screenshot represents the music recognition feature, but GPT-4o misidentified it. This highlights the challenge of accurately interpreting UI elements in an ecosystem of diverse apps with distinct designs. %\C{The detailed prompt is in the appendix}
  }
  \label{fig:qqmusic}
  %\vspace*{-5mm}
  \vspace*{-3mm}
\end{figure}
\footnotetext{\url{https://play.google.com/store/apps/details?id=com.tencent.qqmusic}}

However, the practical deployment of these models faces significant challenges. These challenges stem from the long-tail distribution of app/website variants and their rapid iteration cycles. While core functionalities might appear similar across platforms, critical design divergences exist. For example: shopping cart features in Amazon.com
%\footnote{\url{https://www.amazon.com}} 
and Temu\footnote{\url{https://www.temu.com}} share similarities. In contrast, Pinduoduo\footnote{\url{https://mobile.pinduoduo.com}} (China’s dominant e-commerce platform) eliminates cart functionality entirely. This requires single-item purchases rather than batch checkout. Such inconsistencies extend beyond functionality to interface semantics. 
%When queried about the music recognition button in QQ Music[\C{ Figure~\ref{fig:qqmusic} }], (M)LLMs like GPT-4o often misinterpret its purpose. 
As shown in Figure~\ref{fig:qqmusic}, even advanced MLLMs such as GPT-4o~\citep{hurst2024gpt} can misinterpret the button's actual functionality.
Human users familiar with the app, however, correctly interpret it through learned interaction patterns. Compounding this challenge, apps/websites undergo frequent updates. Amazon Shopping alone released 30 version iterations in 2024\footnote{\url{https://www.apkmirror.com/uploads/?appcategory=amazon-shopping}}. This renders static model knowledge obsolete. Retraining or fine-tuning (M)LLMs for every change proves prohibitively expensive and latency-prone. %In such scenarios, efficient and economical In-Context Learning (ICL) techniques~\citep{dong2024survey} represent a promising direction. These techniques don't require model parameter updates or additional components.

In this paper, we propose \textbf{Autonomous Exploration and Mining of Transition-aware Knowledge for GUI Agent (GUI-explorer)}. 
It synergizes two key components: 
\textbf{(1) Autonomous Exploration of Function-aware Trajectory}. To cover all potential functions of target applications, we design a \textbf{Function-aware Task Goal Generator}. This module automatically constructs function-aware exploration goals by analyzing structural information of the environment, including screenshots and activity lists from APK files. Through systematic exploration, we obtain diverse function-aware trajectories.
\textbf{(2) Unsupervised Mining of Transition-aware Knowledge}. To establish precise operation logic, we develop a \textbf{Transition-aware Knowledge Extractor}. This component extracts effective operation logic through unsupervised analysis of state transitions from structured interaction triples (observation, action, outcome). This eliminates human involvement. Through multimodal state modeling incorporating visual patterns and semantic patterns, the extractor captures operation constraints and outcome dependencies, generating transition-aware knowledge with explicit action-effect correlations.
Finally, by performing visual-semantic retrieval between current screen visuals and the knowledge vector store to construct Dynamic Guidance, it achieves two goals: suppressing the misinterpretation of UI components, and ensuring action proposals align with actual UI states.
This approach facilitates precise, goal-oriented prompt generation. These prompts guide the agent in effectively understanding and interacting with GUI elements. % This approach facilitate ... with GUI elements effectively. 这句话不太确定要不要删掉，这句话能强调这是GUI element级别的Dynamic Guidance
%Figure 1[\C{Need to cite the architecture diagram}] illustrates the architecture of the Autonomous Exploration and Mining of Transition-aware Knowledge for GUI Agent (GUI-explorer).

Our main contributions are listed below:
\begin{itemize}
\item We propose GUI-explorer, a novel training-free agent that integrates two mechanisms: (1) Autonomous exploration of function-aware trajectory through environment-specific structural priors. (2) Unsupervised mining of transition-aware knowledge that extracts atomic screen-operation logic from raw interaction traces. %This dual-phase approach enables continuous knowledge acquisition without model retraining while suppressing element hallucination through visual-semantic retrieval.
\item We conducted comprehensive evaluations of GUI-explorer across AndroidWorld and SPA-Bench benchmarks, our agent achieves 47.4\% and 53.7\% task success rates respectively, outperforming SOTA methods by 2.6\%$\sim$11.7\% improvement. Through ablation studies, we verified that our framework's transition-aware knowledge integration approach reduces prior knowledge errors by 16.0\%.
\item We introduce a benchmark evaluating MLLMs' GUI understanding through 500 curated samples across 43 applications. Results reveal critical limitations in current models (15.2\%$\sim$22.8\% prior knowledge inaccuracies).% and demonstrate our method's effectiveness in bridging these gaps through dynamic knowledge generation (13.4\% error reduction in dynamic comprehension tasks).
\end{itemize}
\begin{figure*}[htb]
  \includegraphics[width=\linewidth]{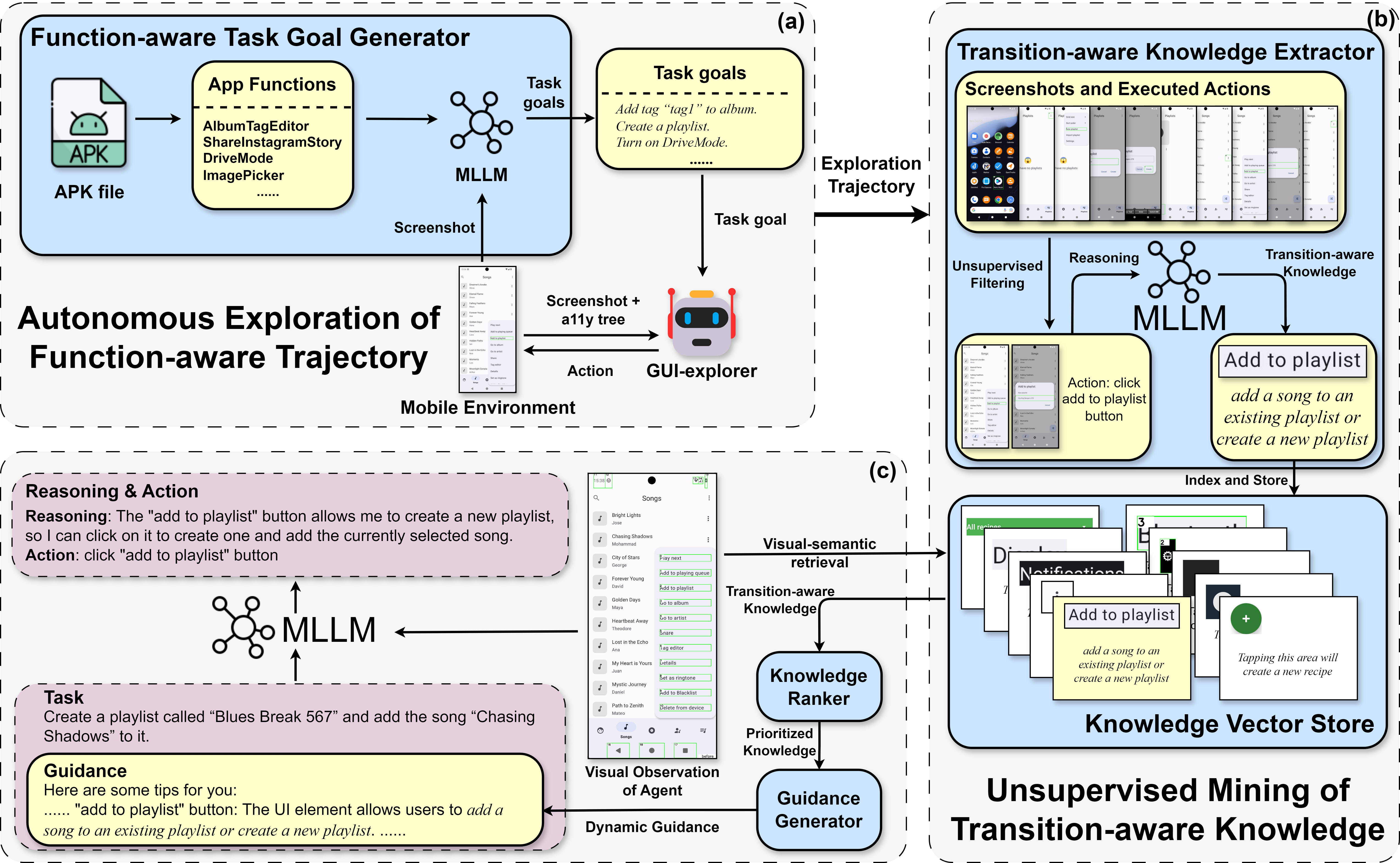}
  \caption{Overview of GUI-explorer. 
(a) Automatically constructing function-aware exploration goals by analyzing structural information from the GUI environment, followed by systematic exploration to collect diverse function-aware trajectories. 
(b) Extracting effective screen-operation logic through unsupervised analysis of structured interaction triples (observation, action, outcome), enabling unsupervised knowledge extraction. 
(c) Performing visual-semantic retrieval between screen visuals and the knowledge vector store to construct Dynamic Guidance achieves dual objectives: preventing UI misinterpretation and ensuring action proposals align with actual UI states.}
  \label{fig:GUI-explorer}
  %\vspace*{-3mm}
  \vspace*{-1mm}
\end{figure*}
\vspace*{-1mm}
\section{Related Work}
\paragraph{GUI Agents} 
%需要训练：CogAgent、Auto-UI、UGround、Aria-UI
%无需训练：SeeAct、MobileAgentV2、M3A
Modern GUI agents leverage MLLMs to interpret interface states and execute actions. %These approaches fall into two categories based on training requirements. \textbf{Training-required agents} typically employ specialized architectures for GUI understanding: CogAgent~\citep{hong2024cogagent} introduces an 18B visual language model with dual-resolution image encoders to recognize fine-grained UI elements, achieving strong performance on Android navigation tasks. Auto-UI~\citep{zhang-zhang-2024-look} bypasses environment parsing through direct pixel-level interaction, while UGround~\citep{gou2024uground} establishes visual grounding models to map referring expressions to GUI coordinates. Aria-UI~\citep{yang2024aria} adopts a pure-vision approach for cross-platform element localization, eliminating reliance on accessibility trees (a11y tree)\footnote{\url{https://developer.mozilla.org/en-US/docs/Glossary/Accessibility_tree}}.  
%\textbf{Training-free agents} primarily utilize in-context learning with general-purpose MLLMs. 
SeeAct~\citep{zheng2024seeact} pioneers GPT-4V~\citep{openai2023gpt4v} for web task automation through visual understanding and HTML-guided action grounding. MobileAgentV2~\citep{NEURIPS2024_0520537b} implements multi-agent collaboration with memory units to track task progress and interface focus. M3A~\citep{rawles2024androidworld} integrates ReAct-style~\citep{yao2023react} reasoning with Set-of-Mark (SoM)~\citep{yang2023set} visual annotations for Android device control, demonstrating zero-shot generalization across applications.  

\paragraph{Exploration \& Knowledge-aware Agents} 
%探索：AppAgent、DigiRL(探索用的task是预先编写好的)、AutoDroid、MobileGPT
%知识利用：CAT、SYNAPSE、ICAL、UI-TARS
Autonomous exploration mechanisms vary in supervision requirements. AppAgent~\citep{10.1145/3706598.3713600} requires manually designed exploration tasks for knowledge acquisition, while AutoDroid~\citep{wen2024autodroid} and MobileGPT~\citep{10.1145/3636534.3690682} generates random action sequences for environment interaction. DigiRL~\citep{zhoudigirl} employs reinforcement learning with Gemini-based~\citep{gemini_2} trajectory filtering to collect successful demonstrations as training data. 

Knowledge utilization strategies focus on experience retention and retrieval. CAT~\citep{feng2024enabling} employs retrieval-augmented generation with task-specific successful trajectories, though limited to pre-collected demonstrations. Synapse~\citep{zheng2023synapse} introduces trajectory-as-exemplar prompting with state abstraction to improve cross-task generalization. ICAL~\citep{sarch2024vlm} abstracts interaction traces into transferable knowledge through visual-language model summarization and human feedback. %UI-TARS~\citep{qin2025ui} establishes long-term memory via online trace bootstrapping and reflection tuning, enabling continuous self-improvement through error correction.  

While existing methods demonstrate progress, four critical limitations persist: (1) Exploration efficiency suffers from random action generation or manual task design; (2) Knowledge extraction relies on successful trajectories or human curation, limiting scalability; (3) Static knowledge bases struggle with rapidly evolving interfaces; (4) Binding knowledge to specific element IDs restricts reuse to identical UIs. %Our work addresses these through autonomous exploration of function-aware trajectory and unsupervised mining of transition-aware knowledge from raw state transitions.

%\vspace*{-1mm}
\section{Autonomous Exploration and Mining of Transition-aware Knowledge for GUI Agent}
\vspace*{-1mm}
%这里主要是用公式把这个方法讲清楚
As illustrated in Figure~\ref{fig:GUI-explorer}, GUI-explorer consists of two main components: autonomous exploration of function-aware trajectory and unsupervised mining of transition-aware knowledge. Building upon the dual components mentioned, we employ visual-semantic retrieval during the agent's task execution to extract relevant knowledge based on the current observation. This retrieval mechanism enables a dynamic knowledge integration process that enhances the agent's decision-making capabilities. Specifically, we construct task-specific guidance by synthesizing the retrieved knowledge with both the current task goal and observational data. This guidance framework facilitates sophisticated reasoning processes, allowing the agent to make more informed decisions while navigating complex task environments.
%\C{To be continued}

\subsection{Autonomous Exploration of Function-aware Trajectory}
\label{sec:Autonomous-Exploration-of-Function-aware-Trajectory}
%注意：这里可以提一下不编写任务仅仅随机操作的方法。和他们的区别在于可以在生成的任务可以作为一个指引不会像随机操作那样在某些页面浪费过度探索步数（例如填表、瀑布流页面（常见于YouTube、推特等app））
The core of our method lies in autonomously generating diverse interaction trajectories without human supervision. This exploration is grounded in environment-specific structural priors. These priors suppress misinterpretations derived from MLLMs' obsolete domain priors. Algorithm~\ref{alg:GUI-Autonomous-Exploration} formalizes this process through two key components. First, anchor-guided task generation leverages interface semantics. Second, depth-first exploration incorporates state restoration mechanisms.

%Given a target environment $E$, we first extract \textit{Exploration Anchors} - structural primitives from $E$'s ground-truth architecture~\appref{sec:Understanding-Exploration-Anchors}.
Given a target environment $E$, we first extract \textit{Exploration Anchors}. These are structural primitives derived from $E$'s ground-truth architecture, as detailed further in Appendix~\ref{sec:Understanding-Exploration-Anchors}. For mobile apps, functional modules declared in manifest files (e.g., "PaymentActivity"). These anchors serve as verifiable constraints during task generation, preventing MLLMs from proposing actions targeting non-existent components. The Task\_Generator function constructs prompts (see Appendix~\ref{prompt:task-goal-generator}) containing current observation $o_t$ and valid anchors, then samples up to $k$ candidate tasks from MLLM outputs. %注意：这里要引用一下放到附录的任务生成的prompt模板

The exploration follows depth-first search (DFS) with configurable branching factor $b$ and depth $d$. This strategy eliminates the first state restoration overhead when expanding child tasks. The elimination occurs because each branch naturally inherits the terminal state of its parent task. This differs from breadth-first search (BFS), which requires resetting to the parent state for each sibling task expansion. Starting from initial state $state_0$, each generated task initiates an exploration branch. After executing a task for up to $s$ steps via Task\_Executor, the environment rolls back to previous state $state_i$. This mechanism enables exhaustive traversal of interface pathways without manual reset. The executor terminates exploration branches under two conditions: when receiving an "END" action, or when reaching maximum steps. This balances thoroughness with computational efficiency.

This design achieves two critical properties: 
(1) \textbf{Semantic Grounding}: Anchors tether generated tasks to actual interface functions. 
(2) \textbf{Quadratic Coverage}: Each $d$-depth exploration with branching factor $b$ yields $O(b^d)$ distinct trajectories, systematically capturing combinatorial interaction patterns.
%\begin{algorithm}[H]
\begin{algorithm}[tb]
\footnotesize
%\scriptsize
\caption{Autonomous Exploration of Function-aware Trajectory}\label{alg:GUI-Autonomous-Exploration}
%\SetAlgoLined %这个会导致 noline 这个设置失效
%\DontPrintSemicolon
\SetKwProg{Fn}{Function}{}{}
\SetKw{KwRequire}{Require}
\SetKw{KwEnsure}{Ensure}
\SetKw{Break}{break}
\SetKw{Return}{return}

\KwIn{Environment $E$, max\_branching\_factor $b$, max\_depth $d$, max\_steps $s$}

\Fn{Explore\_DFS($E, b, d, depth, task, s$)}{
    Task\_Executor($E, task, s$)\;
    
    \If{current\_depth $> d$}{
        \Return\;
    }
    current\_state $\gets$ E.get\_current\_state()\;
    
    child\_tasks $\gets$ Task\_Generator($E, b$)\;
    
    %\ForEach{child\_task in child\_tasks}{
    \For{$i = 0$ \KwTo $length(child\_tasks)-1$}{
    
        \If{$i > 0$}{
            E.restore\_to(current\_state)\;
        }
    
        Explore\_DFS($E, b, d, depth+1, child\_tasks[i], s$)\;
        
    }
}

\BlankLine

\Fn{Task\_Generator($E, k$)}{
    anchors $\gets$ E.app\_functions\;
    
    p $\gets$ ConstructPrompt(E.observation, anchors)\;
    
    \Return{MLLM(p).sample\_top\_k(k)}\;
}

\BlankLine

\Fn{Task\_Executor($E, task, s$)}{
    \For{round = 1 \KwTo s}{
        action $\gets$ MLLM(task, E.observation)\;

        \tcc{We store the observation and action for knowledge vector store construction}
        
        \If{action == "END"}{
            \Return\;
        }
        E.step(action)\;
    }
}

\BlankLine

%E.reset()\;
%
initial\_state $\gets$ E.get\_current\_state()\;

tasks $\gets$ Task\_Generator($E, b$)\;

\ForEach{task in tasks}{
    Explore\_DFS($E, b, d, 0, task, s$)\;
    
    E.restore\_to(initial\_state)\;
}
\end{algorithm}

\subsection{Unsupervised Mining of Transition-aware Knowledge}
%用公式把构建知识库的过程讲清楚。例如探索得到的轨迹信息是 <observation_1, action_1, ..., observation_n, action_n> 使用 <observation_i, action_i, observation_{i+1}> 提取出action_i这个时间步所蕴含的GUI知识。
The knowledge construction process focuses on mining atomic screen-operation logic. These logic are derived from exploration trajectories. Let $\xi = \langle o_1, a_1, ..., o_n, a_n \rangle$ denote an interaction trajectory. This trajectory is collected during autonomous exploration. We extract transition-aware GUI knowledge through a \textbf{Transition-aware Knowledge Extractor} function $F_{extract}$. This function operates on state-action transitions: %最好提一下这个函数借助MLLM来实现，具体实现会在后面的实验章节介绍
\begin{equation}
F_{extract}: (o_i, a_i, o_{i+1}) \rightarrow \{k_i: v_i\}
\end{equation}
where $o_i$ and $o_{i+1}$ represent consecutive observations, $a_i$ denotes the action executed, and $\{k_i: v_i\}$ outputs a set of visual-semantic knowledge entries. Each entry consists of: (1) $k_i$: visual patch of the interacted UI element, (2) $v_i$: operational knowledge (e.g., "Clicking this button opens search history").

%CAT、SYNAPSE、ICAL、UI-TARS
Unlike previous work~\citep{zheng2023synapse,feng2024enabling,sarch2024vlm,qin2025ui}, which requires successful trajectories for in-context learning or fine-tuning, our approach has different requirements. Specifically, we only need valid state transitions. Therefore, we implement a filtering mechanism termed \textbf{Transition Filtering} to filter out invalid state transitions: Discard transitions where $o_i \approx o_{i+1}$. This similarity is measured via perceptual hashing~\citep{marr1980theory}. Such transitions indicate ineffective actions. These occur in two scenarios: when $a_i$ fails to alter the environment (invalid action) or when the environment fails to respond (execution error).

The knowledge vector store $\mathcal{K}$ is structured as a multi-modal index:
\begin{equation}
\mathcal{K} = \bigcup_{\xi \in \Xi} \bigcup_{t=1}^{|\xi|-1} F_{\mathit{extract}}(o_t, a_t, o_{t+1})
\end{equation}
where $\Xi$ denotes all exploration trajectories and $|\xi|$ denotes the total steps of the trajectory $\xi$. %Figure 3\C{[knowledge item visualization]} illustrates the knowledge entry structure, showing UI element patches paired with their operational descriptions.

This knowledge construction process enables \textbf{Continuous Knowledge Refinement}. New explorations iteratively update $\mathcal{K}$ through: 
\begin{equation}
\text{\scriptsize$\displaystyle
\mathcal{K} = \begin{cases} 
\mathcal{K} \setminus \{(k_{\text{old}}, v_{\text{old}})\} \cup \{(k_{\text{old}}, v_{\text{old}} \oplus v_{\text{new}})\} & \text{if } \Phi \\ 
\mathcal{K} \cup \{(k_{\text{new}}, v_{\text{new}})\} & \text{otherwise}
\end{cases}$}
\end{equation}
where $\mathcal{K} \setminus \{(k_{old}, v_{old})\}$ denotes the removal of the original key-value pair from the knowledge vector store, $\oplus$ represents the concatenation of knowledge, condition $\Phi$ is formally defined as:
\begin{equation}
\begin{split}
&\exists (k_{\text{old}},v_{\text{old}}) \in \mathcal{K}  \\ 
&\text{s.t.} \quad \left\{\begin{array}{ll}
\cos\big(\text{\textit{Emb}}(k_{\text{new}}), \text{\textit{Emb}}(k_{\text{old}})\big) \geq \delta_k \\
\cos\big(\text{\textit{Emb}}(v_{\text{new}}), \text{\textit{Emb}}(v_{\text{old}})\big) \leq \delta_v
\end{array}\right.
\end{split}
\end{equation}
where $\delta_k$ and $\delta_v$ are similarity thresholds for key matching ($\geq 0.99$) and value merging ($\leq 0.1$) respectively, $\cos(\cdot)$ is cosine similarity, and $\text{Emb}(\cdot)$ is the embedding function. This prevents redundant entries while capturing novel interface behaviors. 

\begin{figure}[tb]
  \includegraphics[width=\linewidth]{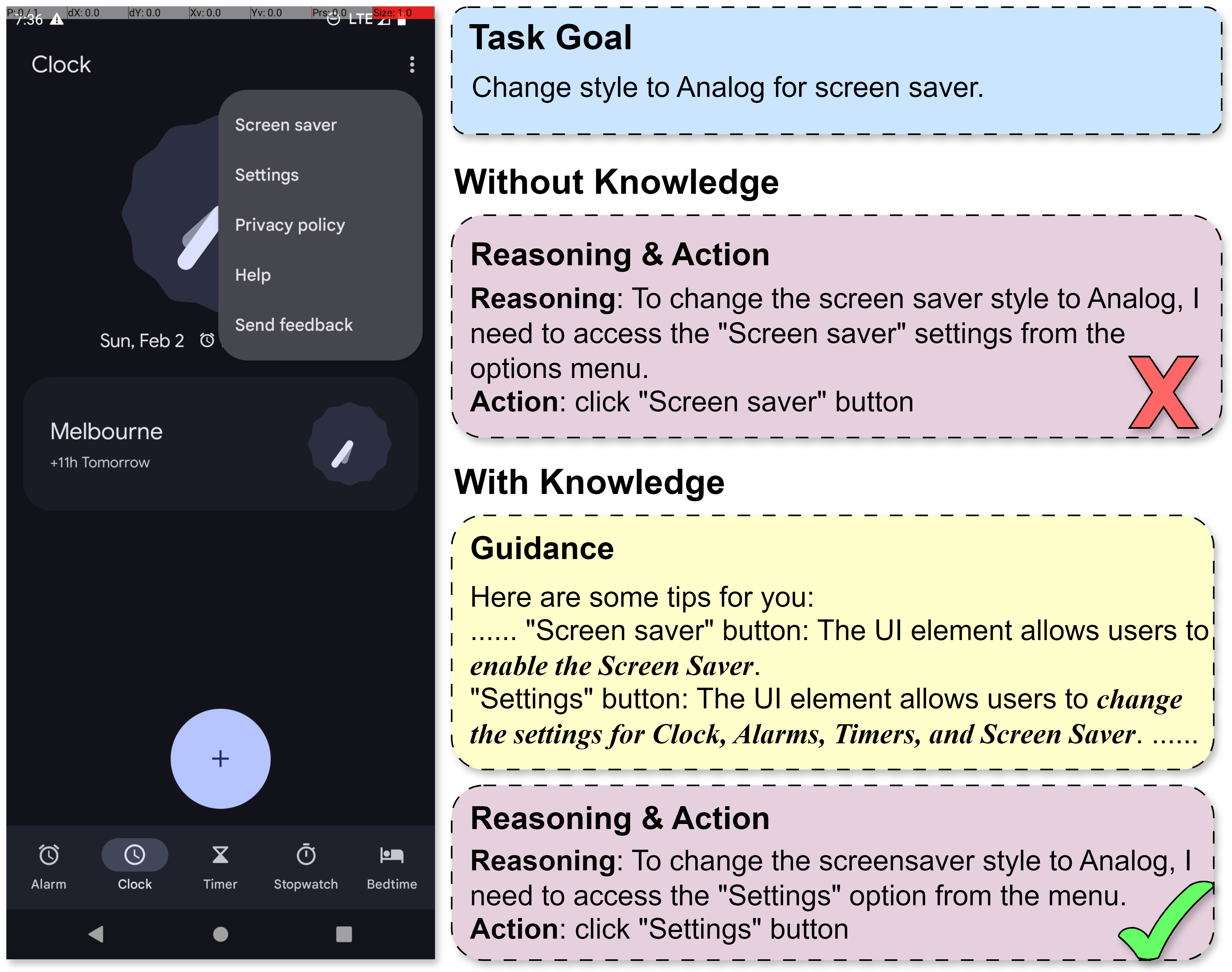}
  \caption{Without transition-aware knowledge as reliable prior information, MLLMs may fail to reason correctly due to outdated prior knowledge or diverse GUI designs.}
  \label{fig:compare}
  %\vspace*{-5mm}
\end{figure}

Figure~\ref{fig:compare} demonstrates the importance of transition-aware knowledge.
%Our experiments in Section~\ref{transition-aware-knowledge-accuracy} show this pipeline achieves 86.6\% knowledge accuracy compared to human annotations.
\subsection{Dynamic Guidance for GUI Agent}
%或许这里需要训练一个简单的rank模型（可以用小模型，例如0.5b的那种纯文本模型即可）将检索结果进行rank（不再按照som的数字编号从小到大，而是将检索出的文本和task description用模型进行判断并排序）。训练模型的事情可以在后面的章节讲，这里主要是用公式表达出这个模型是怎么工作的。或者用train-free的方式来实现这个ranker会更好（具体思路：输入task description和两个候选项，让LLM选择哪个候选项更适合task。这样只需要重载小于号就能用归并排序高效地对整个候选列表进行task相关度排序）
%这里用伪代码来介绍 Dynamic Guidance ，然后用公式来介绍rank模型
%\begin{algorithm}[H]
\begin{algorithm}[htb]
\footnotesize
\caption{Dynamic Guidance for GUI Agent}\label{alg:knowledge-guided-execution}
\SetKwProg{Fn}{Function}{}{}
\SetKw{KwRequire}{Require}
\SetKw{KwEnsure}{Ensure}
\SetKw{Break}{break}
\SetKw{Return}{return}

\KwIn{Environment $E$, Instruction $I$, Knowledge\_Vector\_Store $\mathcal{K}$, Knowledge\_Ranker $Ranker$, max\_steps $s$}

\Fn{Get\_Guidance(obs, $I$, $\mathcal{K}$)}{

    annot\_scr $\gets$ Get\_Annotated\_Screenshot(obs)\;
    
    ui\_elements $\gets$ Extract\_UI\_Elements(obs)\;
    
    all\_knol $\gets \emptyset$ \tcp*{all\_knowledge}
    
    \ForEach{ui\_element \textbf{in} ui\_elements}{
    
        all\_knol.append(Retrieve\_Knowledge($\mathcal{K}$, ui\_element))\;
        
    }
    
    prioritized\_knol $\gets$ $Ranker(I, all\_knol)$\;
    
    guidance $\gets$ Create\_Guidance\_Prompt($I$, rioritized\_knol, annot\_scr)\;
    
    \Return{guidance}\;
}

\BlankLine

%E.reset()\;
%
\For{$idx = 1$ \KwTo $s$}{

    obs $\gets$ E.observation\;
    
    operational\_guid $\gets$ Get\_Guidance(obs, $I$, $\mathcal{K}$)\;
    
    action $\gets$ MLLM($I$, operational\_guid, obs)\;
    
    \If{action == "END"}{
        \Break\;
    }
    E.step(action)\;
}
\end{algorithm}
The dynamic guidance mechanism connects acquired Transition-aware Knowledge to real-time task execution. This connection is achieved through a ranking architecture. As detailed in Algorithm~\ref{alg:knowledge-guided-execution}, our approach uses a two-phase process. The first phase involves visual-semantic knowledge retrieval. The second phase performs instruction-aware prioritization.

\paragraph{Knowledge Ranking Formulation}
Given an instruction $I$ and candidate knowledge entries $\mathcal{C} = \{k_1,...,k_n\}$, we define the optimal knowledge ordering $\mathcal{C}^*$ through pairwise utility comparison:
\begin{equation}
\text{\footnotesize$\displaystyle
\mathcal{C}^* = \argmax_{\pi \in \Pi(\mathcal{C})} \sum_{i=1}^{|\mathcal{C}|-1} {int}(u(k_{\pi(i)}, I) \geq u(k_{\pi(i+1)}, I))$}
\end{equation}
where $\Pi(\mathcal{C})$ denotes all permutations of $\mathcal{C}$, $int(\cdot)$ converts bool to integer (false as 0, true as 1), and utility function $u(k,I)$ measures the relevance between knowledge entry $k$ and instruction $I$. We implement $u(\cdot)$ through an MLLM-based pairwise comparator:
\begin{equation}
\text{\footnotesize$\displaystyle
u(k_a, I) > u(k_b, I) \Leftrightarrow f_{\text{rank}}(g(I, k_a, k_b)) = 1$}
\end{equation}
where $g(\cdot)$ constructs the ranking prompt (see Appendix~\ref{prompt:knowledge-ranker}), and $f_{\text{rank}}$ represents the MLLM's binary classification. When the classification result is 1, it indicates $k_a$ is more helpful than $k_b$ for this instruction. When the result is 2, it means $k_b$ is more helpful than $k_a$. This formulation enables efficient sorting through a modified merge sort algorithm:
\begin{equation}
\text{\scriptsize$\displaystyle
\text{Sort}(\mathcal{C}, I) = \begin{cases} 
\mathcal{C} & |\mathcal{C}| \leq 1 \\
\text{Merge}(\text{Sort}(\mathcal{C}_L, I), \text{Sort}(\mathcal{C}_R, I), I) & \text{otherwise}
\end{cases}$}
\end{equation}
The merge operation recursively compares head elements from sorted sublists using $f_{\text{rank}}$:
\begin{equation}
\text{\scriptsize$\displaystyle
\text{Merge}(A, B, I) = \begin{cases} 
[a_0] \oplus \text{Merge}(A_{1:}, B, I) & f_{\text{rank}}(g(I, a_0, b_0)) = 1 \\
A \oplus B & A = \emptyset \lor B = \emptyset \\
[b_0] \oplus \text{Merge}(A, B_{1:}, I) & \text{otherwise}
\end{cases}$}
\end{equation}
where $a_0$ and $b_0$ denote the first elements of lists $A$ and $B$ respectively.

\paragraph{Operational Guidance Generation}
At each execution step $t$, the system: 
(1) Extracts UI elements $\mathcal{U}_t$ from current observation $o_t$;
(2) Retrieves associated knowledge entries $\mathcal{K}_t \subseteq \mathcal{K}$;
(3) Sorts entries via $\mathcal{K}_t^* = \text{Sort}(\mathcal{K}_t, I)$;
(4) Constructs guidance prompt $p_t$ with relevant knowledge.

%Our experiments demonstrate this approach achieves \C{A\%} higher task success compared to baselines \C{B\%} on AndroidWorld. The pairwise ranking strategy reduces hallucination errors by \C{C\% on GUI-KRB task type 1} by suppressing irrelevant historical knowledge. 

%将这个图片放到附录，记得在这里说一下是在附录。而且也要改一下表述。例如：在附录xxx展示了一个dynamic guidance mechanism enables precise alignment between operational knowledge and real-time interface states的例子
%As shown in Figure 4\C{M3A example (without M3A)}, the dynamic guidance mechanism enables precise alignment between operational knowledge and real-time interface states.
As shown in Figure~\ref{fig:GUI-explorer}~(c), the dynamic guidance mechanism enables precise alignment between operational knowledge and real-time interface states.

%\C{To be continued}

\section{GUI-Knowledge Reasoning Benchmark}
We introduce the GUI-Knowledge Reasoning Benchmark (\textbf{GUI-KRB}). This benchmark evaluates MLLMs’ accuracy in two areas: prior knowledge accuracy and dynamic UI comprehension for mobile environments. Existing benchmarks primarily focus on task completion. In contrast, GUI-KRB assesses models’ fundamental understanding of UI elements and their behaviors. It contains 500 carefully curated samples spanning 43 applications across 8 categories. Appendix~\ref{sec:GUI-KRB-Distributions} shows the proportion of apps in each category.

\subsection{Tasks and Metrics}
%\paragraph{Tasks and Metrics}
GUI-KRB includes two evaluation tasks:
(1) \textbf{Prior Knowledge Assessment}: Models must identify the functionality of specified UI elements. They are given a single screenshot, its accessibility tree, and a task context about this element. This task simulates the planning phase in GUI automation. During planning, agents must understand element functionality before acting. Success here indicates effective use of prior training knowledge.
(2) \textbf{Dynamic Comprehension Assessment}: Models analyze UI element functionality by comparing pre-interaction and post-interaction states within the task context of this transition. These states include screenshots and accessibility trees. This task evaluates reasoning about cause-effect logic in GUI interactions. It simulates the knowledge extraction method we use in this paper.

For both tasks, responses are evaluated against human-annotated keywords. A response is considered correct if it contains at least 50\% of expert-identified keywords. This metric balances precision with flexibility for valid phrasings. (During keyword labeling, we include up to 50\% synonyms to accommodate diverse responses.)

\subsection{Annotation Process}
%\paragraph{Annotation Process}
GUI-KRB was built through a rigorous multi-stage process:
(1) \textbf{Trajectory Collection}: We collected over 300 task execution trajectories in a mobile environment. These trajectories contain more than 7,000 interaction steps across diverse mobile applications. They capture authentic user interactions in real-world scenarios.
(2) \textbf{Element Extraction}: From these trajectories, we extracted individual UI elements using bounding box information from accessibility trees. To ensure diversity and remove redundancy, we eliminated duplicate elements using perceptual hashing techniques~\citep{marr1980theory}.
(3) \textbf{Keyword Annotation}: Human experts identified essential keywords uniquely associated with each UI element's functionality. These keywords capture both the element's immediate purpose and its broader role in the interface.
(4) \textbf{Validation}: The authors conducted a comprehensive review of all annotations, verifying keyword accuracy and ensuring consistent annotation quality across the dataset.

The final dataset provides triplets of target UI elements, their corresponding screen states (before and after interaction), and expert-validated keyword sets. Example annotations are provided in Appendix~\ref{sec:GUI-KRB-sample}.

\section{Experiments}
%这里简要介绍使用的指标、数据集、模型。并且讲一下一些实现上的技术细节
\subsection{Experimental Setup}
%\subsubsection{Metrics} 这个小节或许合并到\subsubsection{Datasets}中会更好
%介绍任务成功率、总步数、以及GUI-Knowledge Reasoning Benchmark (GUI-KRB)里面的指标
%\C{To be continued}

\subsubsection{Datasets}
%介绍评测的GUI benchmark，以及（简要介绍如何构建以及分布（包含哪些类别的app各个类别有多少条/占比多少））GUI-Knowledge Reasoning Benchmark (GUI-KRB)以及它的用途/目的

We evaluate GUI-explorer on two comprehensive, open-source benchmarks: MIT-licensed SPA-Bench~\citep{chen2024spa} and Apache-2.0-licensed AndroidWorld~\citep{rawles2024androidworld}. Both benchmarks emphasize real-world GUI and provide automated evaluation pipelines for rigorous agent assessment.

%注：SPA-Bench这里需要提及中文app以及cross app的任务吗？如果这里提了，后面又没有评测，似乎有点说不通。现在这样子只介绍到我们用到的部分可以吗？
\paragraph{SPA-Bench}
%\textbf{SPA-Bench} 
SPA-Bench 
is a benchmark simulating daily smartphone usage scenarios with 58 mainstream apps  (e.g., Facebook and Gmail). It contains three progressively challenging task levels (Level 1-3), where Level 3 represents the most complex real-world workflows. %Key features include:
%\textbf{Daily Apps}: Focuses on high-frequency applications mirroring authentic user behavior patterns;
%\textbf{Automated Evaluation Pipeline}: Implements a coarse-to-fine verification system combining OCR-based component matching and MLLM semantic validation, achieving 90.5\% F1-score for task success detection;
%\textbf{Multi-Dimensional Metrics}: Tracks success rate, step efficiency ratio (compared to human demonstrations), API cost, execution latency, etc.

\paragraph{AndroidWorld}
%\textbf{AndroidWorld} 
AndroidWorld 
is an Android environment featuring 116 tasks across 20 real-world apps. The benchmark dynamically generates task variants through randomized parameters (e.g., message content, contact names, and calendar dates), creating millions of unique task instantiations. %Key characteristics include:
%\textbf{Real-World Apps}: Covers system utilities (Settings, Contacts) and open-source apps (Simple Calendar Pro, Markor) with native Android UI patterns;
%\textbf{State-Based Validation}: Verifies task success through Android system state inspection (e.g., SQLite database entries, file system changes) rather than superficial UI matching;
%\textbf{Composite Tasks}: Includes multi-step workflows requiring inter-app coordination (e.g., "Create calendar event → Share details via SMS"). The evaluation reports task success rate (SR).

%\C{To be continued}

\subsubsection{Implementation Details}
%prompt模板放到appendix中，这里主要讲选用什么模型生成知识以及怎么构建prompt让MLLM生成知识。以及如何进行in-context learning（放PPT中画好的那个示意图即可）%探索的超参数、检索的超参数介绍一下，然后说一下一共获得的知识数量。最好用一个表格或者饼图（推荐饼图）展示各个类别的app的知识数量

To ensure fair evaluation across benchmarks, we carefully selected base models according to their characteristics. For SPA-Bench and AndroidWorld, we adopted GPT-4o~\citep{hurst2024gpt} as the unified base model, which has been the de facto standard model in prior works including, but not limited to, SPA-Bench~\citep{chen2024spa} and Aria-UI~\citep{yang2024aria}, eliminating performance variance caused by heterogeneous model capabilities. In contrast, for GUI-KRB, we intentionally utilized the weakest-performing Qwen2-VL-72B-Instruct-GPTQ-Int4~\citep{Qwen2-VL} as our base model, to rigorously validate the robustness of our method. 

We configured the exploration process with a branching factor of 10, a maximum depth of 5, and a step limit of 30 for AndroidWorld and SPA-Bench. This setup facilitated the automated discovery of over 1,300 knowledge items (detailed distribution in 
%Figure~\ref{fig:distribution_of_knowledge_and_number_of_apps}
\appref{sec:Distribution-of-Transition-aware-Knowledge}) across 46 applications. For visual-semantic retrieval, we utilized google/siglip-so400m-patch14-384\footnote{\url{https://huggingface.co/google/siglip-so400m-patch14-384}} as the embedding model. Hardware configurations are provided in Appendix~\ref{sec:Hardware-configurations}.

%Hardware configurations were optimized for cost-effectiveness: Most experiments ran on a single NVIDIA GeForce RTX 4070 Laptop GPU (8GB VRAM). For GUI-KRB evaluations involving open-source MLLMs, we scaled to two NVIDIA L40S GPUs (48GB VRAM) to accommodate larger VRAM requirements. 
\subsubsection{Comparative Baselines}
%介绍一下那些benchmark中出现过的模型，baseline就是没有使用GUI-explorer的模型，他们就是baseline，用他们和加了GUI-explorer的版本进行对比。类似于 https://arxiv.org/pdf/2310.05007 这样做，我们用现有的实验结果即可，然后我们可以把 GUI-explorer 加到两个本来就挺好的模型上面（目前已经加到了M3A上面，或许再选一个纯视觉的Aria-UI？）
%简要介绍一些baseline
%写法2（推荐用这个）
%我们选取了AppAgent、AutoDroid（AppAgent和AutoDroid都拥有探索和纯文本知识提取，AppAgent需要人工编写探索任务，而AutoDroid通过随机生成action去操作GUI来获取探索轨迹来实现无需探索任务的探索）、DigiRL（DigiRL通过强化学习的方式在环境中探索并且收集训练数据，它使用Gemini（引用到Gemini）过滤出成功的轨迹作为训练数据）作为baseline。
%与此同时，我们还汇报了各个benchmark上其他方法的结果作为性能的参考。
%注意：DigiRL探索用的task是预先编写好的，在 https://github.com/DigiRL-agent/digirl/issues/20 可以得知会使用 https://github.com/DigiRL-agent/digirl/tree/master/digirl/environment/android/assets/task_set 这里的task进行探索并且获得探索轨迹用于训练（文中的强化学习应该指的是把正确和错误（用geemini进行判断）的轨迹都用上来进行RL）
We select three baselines with exploration and knowledge extraction capabilities for comprehensive comparison. AppAgent~\citep{10.1145/3706598.3713600} requires manually designed exploration tasks to guide its interaction with GUI environments for knowledge acquisition, whereas AutoDroid~\citep{wen2024autodroid} eliminates task-specific human effort by autonomously generating random action sequences to collect exploration trajectories. Both methods extract structured text-based knowledge from raw textual observations during exploration. DigiRL~\citep{zhoudigirl} adopts a distinct reinforcement learning framework to iteratively explore environments while utilizing the Gemini~\citep{gemini_2} model to filter successful trajectories as training data, enabling adaptive exploration with minimal human intervention. For completeness, we also report results from additional baselines in their respective benchmark papers as performance references. %All baseline implementations adhere to their original design specifications for fair comparison.

\subsection{Experimental Results}
\begin{table}[tb]%\small
%\vspace*{-3mm}
\centering
%\scriptsize
\resizebox{\columnwidth}{!}{
\begin{tabular}{lllc}
\toprule[1.5pt]
\textbf{Agent}         & \textbf{Input}              & \textbf{Base Model} & {\begin{tabular}[c]{@{}c@{}}\textbf{Task Success} \\ \textbf{Rate (\%)}\end{tabular}} \\
\midrule
AppAgent~\citep{10.1145/3706598.3713600}      & SoM   & GPT-4o     & 14.0 \\
AutoDroid~\citep{wen2024autodroid}     & a11y tree              & GPT-4o     & 12.0 \\
%Auto-UI~\citep{zhang-zhang-2024-look}      & screen             & Auto-UI    & 0 \\
CogAgent~\citep{hong2024cogagent}     & screen             & CogAgent   & 0 \\
DigiRL~\citep{zhoudigirl}       & screen             & DigiRL     & 0 \\
M3A~\citep{rawles2024androidworld}           & SoM   & GPT-4o     & 42.0 \\
MobileAgentV2~\citep{NEURIPS2024_0520537b} & SoM   & GPT-4o     & 20.0 \\
SeeAct~\citep{zheng2024seeact}        & SoM   & GPT-4o     & 12.0 \\
%T3A~\citep{rawles2024androidworld}          & a11y tree              & GPT-4o     & 26.0 \\
\rowcolor[HTML]{E7EEFE}
GUI-explorer (Ours)     & SoM   & GPT-4o     & \textbf{53.7} \\
\bottomrule[1.5pt]
\end{tabular}
}
\caption{Performance comparison on SPA-Bench single-app English Level 3 tasks. Results for the first 7 agents are from the SPA-Bench~\citep{chen2024spa}. SoM~\citep{yang2023set} utilizes the bounding boxes (bbox) recorded in the a11y tree to annotate UI elements with numerical labels in screenshots.}
\label{tab:results-eng-single-lv3-spa}
%\vspace*{-3mm}
\end{table}
Our comprehensive evaluation demonstrates GUI-explorer's superior performance across multiple dimensions. As shown in Table~\ref{tab:results-eng-single-lv3-spa}, GUI-explorer achieves 53.7\% task success rate on SPA-Bench single-app English Level 3 tasks. This represents a 28.1\% absolute improvement over M3A, the previous state-of-the-art. Our transition-aware knowledge mining approach proves highly effective in complex, real-world scenarios.

The AndroidWorld results in Table~\ref{tab:results-aw} further validate GUI-explorer's generalizability. Our agent achieves 47.4\% success rate. This surpasses vision-centric Aria-UI at 44.8\%. It also outperforms multimodal M3A at 40.5\%.

The GUI-KRB evaluation reveals critical insights about MLLMs' GUI reasoning limitations. GPT-4o shows an 18.2\% prior knowledge error rate. These errors mainly stem from the misinterpreting of UI components and outdated interface understanding. Our method reduces these errors by 16.0\% when applied to Qwen2-VL-72B-Instruct-GPTQ-Int4. This demonstrates the effectiveness of transition-aware knowledge. The dynamic comprehension assessment shows similar improvements. GUI-explorer-enabled models achieve 13.4\% lower error rates than base models.

\subsection{Analysis and Discussion}
%消融实验。例如检索的top k超参数、知识库的泛化性测试、不同MLLM根据两张屏幕截图生成的知识的准确度（这里就是唯2要用到 GUI-Knowledge Reasoning Benchmark (GUI-KRB) 的地方了）
\begin{table*}[htb]\small
\centering
%\scriptsize
%\footnotesize
\resizebox{\linewidth}{!}{
\begin{tabular}{lccccccc}
\toprule[1.5pt]
\textbf{App Category} & \textbf{\begin{tabular}[c]{@{}c@{}}Retrieval Time \\ per Step (sec)\end{tabular}} & \textbf{\begin{tabular}[c]{@{}c@{}}Ranking Time \\ per Step (sec)\end{tabular}} & \textbf{\begin{tabular}[c]{@{}c@{}}Reasoning Time \\ per Step (sec)\end{tabular}} & \textbf{\begin{tabular}[c]{@{}c@{}}Total Time \\ per Step (sec)\end{tabular}} & \textbf{\begin{tabular}[c]{@{}c@{}}Ranking Cost per \\ Step ($0.1$USD)\end{tabular}} & \textbf{\begin{tabular}[c]{@{}c@{}}Reasoning Cost \\ per Step (USD)\end{tabular}} & \textbf{\begin{tabular}[c]{@{}c@{}}Total Cost per \\  Step (USD)\end{tabular}} \\
\midrule
Travel \& Navigation    & 7.663 & \textbf{33.084} & 31.400 & 72.147 & \textbf{0.017} & 0.066 & 0.068 \\
Shopping \& Finance     & \textbf{8.613} & 24.922 & \textbf{36.622} & 70.157 & 0.013 & 0.063 & 0.065 \\
News \& Reading         & 8.123 & 17.317 & 29.272 & 54.712 & 0.008 & 0.053 & 0.053 \\
System Applications     & 6.955 & 31.083 & 34.513 & \textbf{72.552} & 0.016 & 0.065 & 0.067 \\
Productivity \& Tools   & 7.136 & 28.091 & 28.382 & 63.609 & 0.016 & 0.064 & 0.066 \\
Media \& Entertainment  & 7.549 & 32.481 & 30.586 & 70.615 & 0.017 & \textbf{0.066} & \textbf{0.068} \\
Communication \& Social & \textbf{6.176} & 25.662 & \textbf{27.293} & 59.130 & 0.013 & 0.057 & 0.058 \\
Food \& Lifestyle       & 6.304 & \textbf{9.511}  & 30.481 & \textbf{46.296} & \textbf{0.004} & \textbf{0.041} & \textbf{0.042} \\
\midrule
Overall                 & 7.120 & 28.462 & 30.796 & 66.378 & 0.015 & 0.062 & 0.064 \\
\bottomrule[1.5pt]
\end{tabular}
}
\caption{Per-Step Computational Overhead Analysis: Breakdown of time consumption (seconds) and API costs (USD) across application categories. Note that Ranking Cost per Step is presented in Dimes (0.1 USD) for better readability due to its small magnitude.
%Reasoning components represent baseline agent expenditures, while retrieval and ranking reflect GUI-explorer's incremental overhead. 
}
\label{tab:analysis-time-cost}
%\vspace*{-5mm}
\end{table*}

\begin{table}[htb]%\small
\centering
%\scriptsize
\resizebox{\columnwidth}{!}{
\begin{tabular}{lllc}
\toprule[1.5pt]
\textbf{Agent}         & \textbf{Input}              & \textbf{Base Model} & {\begin{tabular}[c]{@{}c@{}}\textbf{Task Success} \\ \textbf{Rate (\%)}\end{tabular}} \\
\midrule
\textit{Human~\citep{rawles2024androidworld}}            & \textit{screen}             & \textit{-}     & \textit{80.0} \\
%T3A~\citep{rawles2024androidworld}                  & a11y tree              & GPT-4-turbo & 30.6 \\
Aguvis~\citep{xu2024aguvis}              & screen             & GPT-4o     & 37.1 \\
AppAgent~\citep{10.1145/3706598.3713600}            & SoM             & GPT-4o     & 14.9 \\
Aria-UI~\citep{yang2024aria}             & screen             & GPT-4o     & 44.8 \\
AutoDroid~\citep{wen2024autodroid}     & a11y tree              & GPT-4o     & 15.7 \\
DigiRL~\citep{zhoudigirl}       & screen             & DigiRL     & 0.9 \\
M3A~\citep{rawles2024androidworld}                 & SoM   & GPT-4o     & 40.5 \\
Ponder\&Press~\citep{wang2024ponder}     & screen             & GPT-4o     & 34.5 \\
SeeAct~\citep{rawles2024androidworld}               & SoM   & GPT-4-turbo & 15.5 \\
UGround~\citep{gou2024uground}             & screen             & GPT-4o     & 32.8 \\
\rowcolor[HTML]{E7EEFE}
GUI-explorer (Ours) & SoM & GPT-4o     & \textbf{47.4} \\
\bottomrule[1.5pt]
\end{tabular}
}
\caption{Performance comparison on AndroidWorld. }%Human、SeeAct、T3A的结果来自\C{[relevant citation]}，UGround的结果来自\C{[relevant citation]}，Ponder \& Press的结果来自\C{[relevant citation]}，Aguvis的结果来自\C{[relevant citation]}，Aria-UI的结果来自\C{[relevant citation]}
\label{tab:results-aw}
%\vspace*{-3mm}
\end{table}

\begin{table}[htb]%\small
\centering
%\footnotesize
\resizebox{\columnwidth}{!}{
\begin{tabular}{lcc}
\toprule[1.5pt]
\textbf{Model} & \textbf{\begin{tabular}[c]{@{}c@{}}Prior Knowledge \\ Error Rate (\%)\end{tabular}} & \textbf{\begin{tabular}[c]{@{}c@{}}Dynamic Comprehen- \\ sion Rrror Rate (\%)\end{tabular}}\\
\midrule
Qwen2-VL~\citep{Qwen2-VL} & 22.8 & 19.8 \\
Qwen2.5-VL~\citep{Qwen2.5-VL} & 16.6 & 14.0 \\
%UI-TARS~\citep{qin2025ui} & 18.0 & 14.2 \\
Gemini 2.0 Flash~\citep{gemini_2} & 15.2 & 11.2 \\
GPT-4o~\citep{hurst2024gpt} & 18.2 & 13.4 \\
%\midrule
\rowcolor[HTML]{E7EEFE}
GUI-explorer (w/o Ranker) & 9.8 & 6.8 \\
\rowcolor[HTML]{E7EEFE}
GUI-explorer & \textbf{6.8} & \textbf{6.4} \\
\bottomrule[1.5pt]
\end{tabular}
}
\caption{Performance comparison on GUI-KRB. For all methods, we selected the highest-performing models within device VRAM constraints: Qwen2-VL-72B-Instruct-GPTQ-Int4 for Qwen2-VL, and Qwen2.5-VL-7B-Instruct for Qwen2.5-VL. }%在本实验中，在设备VRAM允许的范围内均选择了对应方法性能最强的模型。对于Qwen2-VL选择了Qwen2-VL-72B-Instruct-GPTQ-Int4，对于Qwen2.5-VL选择了Qwen2.5-VL-7B-Instruct，对于UI-TARS选择了UI-TARS-7B-DPO。
\label{tab:results-gui-krb}
%\vspace*{-6mm}
\end{table}
\begin{figure}[htb]
%\vspace{-5mm}
  \includegraphics[width=\linewidth]{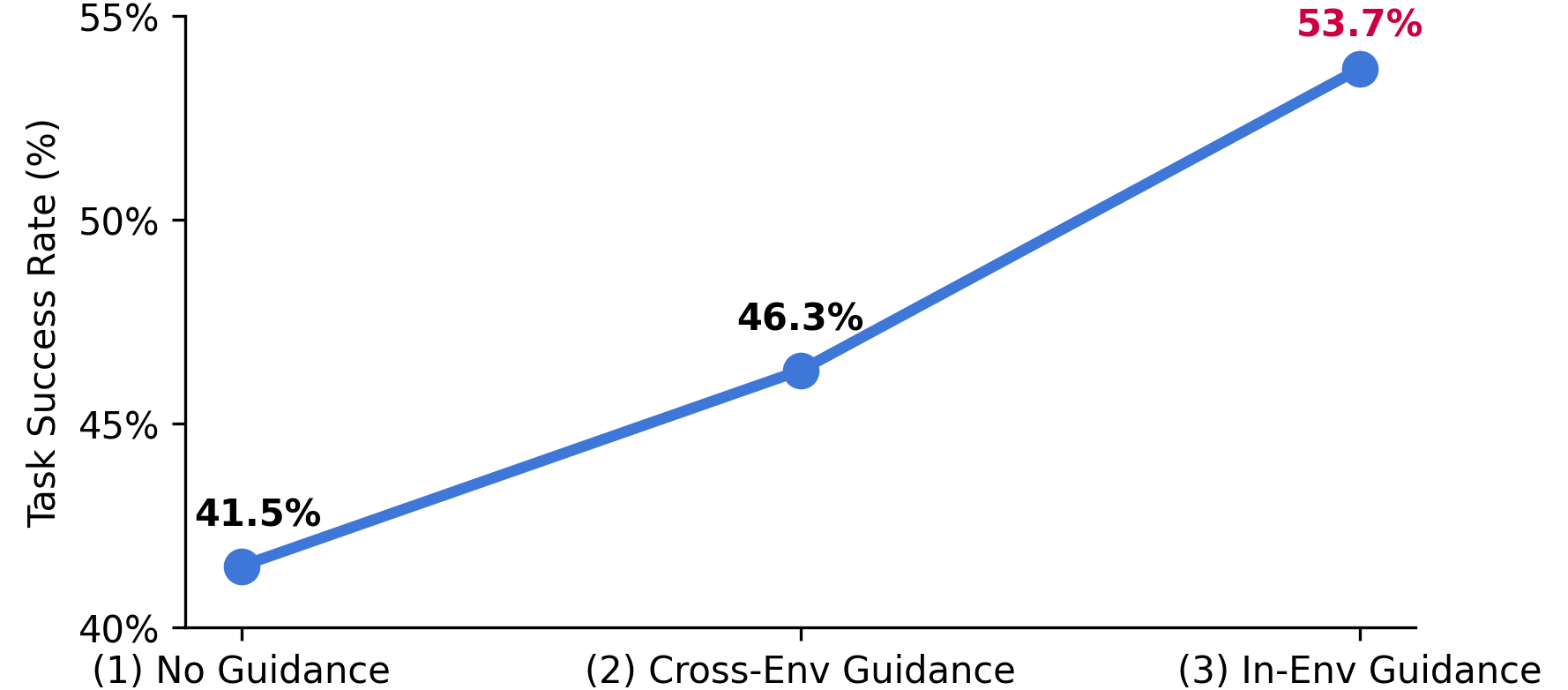}
  \caption{Ablation study of operational guidance configurations on SPA-Bench:
(1) Baseline without dynamic guidance,
(2) Guidance derived from cross-environment exploration (AndroidWorld),
(3) Guidance generated through in-environment exploration (SPA-Bench).}
  \label{fig:ablation_study_on_operational_guidance}
  %\vspace*{-5mm}
\end{figure}

The ablation study in Figure~\ref{fig:ablation_study_on_operational_guidance} quantifies the impact of our key components. Removing dynamic guidance construct by transition-aware knowledge causes a 12.2\% performance drop. This emphasizes the critical role of transition-aware knowledge. Cross-Environment Guidance improves performance by 4.3\% compared to No Guidance. This demonstrates that our transition-aware knowledge exhibits promising generalization capabilities. It effectively guides agent reasoning even in previously unseen scenarios. The knowledge learned can transfer across different UI environments.

Our computational overhead analysis appears in Table~\ref{tab:analysis-time-cost}. It reveals practical tradeoffs. The ranking component contributes 42.9\% of time. This comes primarily from MLLM-based pairwise comparisons. However, we use a merge sort implementation. This ensures $O(n \log n)$ complexity. This keeps practical costs acceptable (0.0015 USD/step average). Additionally, Table~\ref{tab:results-gui-krb} shows another benefit. The ranking component reduced the error rate by 3\% by prioritizing more relevant knowledge.

The GUI-KRB results expose two fundamental limitations in current MLLMs. First, there are persistent prior knowledge gaps. Even Gemini 2.0 Flash~\citep{gemini_2} has a 15.2\% error rate. Second, there is limited dynamic reasoning capability. %GUI-explorer addresses both issues through transition-aware knowledge. The transition-aware knowledge bypasses outdated priors. 

\label{transition-aware-knowledge-accuracy}
%Since Dynamic Comprehension task in GUI-KRB and transition-aware knowledge mining are basically the same task. GPT-4o achieved an 86.6\% correct rate in Dynamic Comprehension. This indicates that knowledge in the Knowledge Vector Store we constructed with GPT-4o is also about 86.6\% correct. 
The GUI-KRB Dynamic Comprehension task, equivalent to transition-aware knowledge mining, achieved 86.6\% accuracy with GPT-4o, indicating comparable reliability in our GPT-4o-built Knowledge Vector Store.

\section{Conclusion}
We present GUI-explorer, a 
%training-free 
GUI agent designed to address two key challenges: misinterpretation of UI components and knowledge obsolescence. Our approach achieves this through autonomous exploration and transition-aware knowledge mining. Experimental results demonstrate our SOTA performance across major benchmarks.%, achieving 53.7\% success rate (SR) on SPA-Bench and 47.4\% SR on AndroidWorld. The method reduces Prior Knowledge errors by 16.0\% compared to existing approaches. 
We introduce the GUI-KRB benchmark, which reveals fundamental limitations in current MLLMs' interface understanding capabilities. Our dynamic guidance mechanism effectively mitigates these limitations. %Future research will focus on optimizing knowledge retrieval efficiency. GUI-explorer advances practical GUI automation in dynamic digital ecosystems through its self-evolving architecture.

\section*{Limitations}
%目前只找到并且验证了适用于mobile场景的Exploration Anchors（For mobile apps, functional modules declared in manifest files (e.g., "PaymentActivity").）。对于web场景以及desktop场景仍然需要寻找合适的信息作为Exploration Anchors来实现无监督的Function-aware Task Goal Generator。也许对于web场景来说，sitemap会是一个很好的Exploration Anchor（For websites, this corresponds to sitemap nodes (e.g., "/product/*" URLs);），但是我们尚未在该领域进一步研究。
%Knowledge Ranker的速度还是太慢了，目前排序占据了将近一半的耗时。在实验过程中我们观察到这和OpenAI的API速率限制有关系，在我们的观察中，每秒钟只完成了几次API调用。我们也尝试了将开源模型（参数量为0.5b的LLM）部署在本地作为Ranker的base model，但是受限于设备的性能或者网络传输的开销，速度并没有明显提升，考虑到小模型的Reasoning能力比较弱，并不值得用少量的速度提升换取性能的下降，所以我们就没有继续进行这个实验了。或许我们应该训练一个端到端的小模型作为Ranker，以此来兼顾Ranker的速度和性能。

While GUI-explorer demonstrates significant advancements in GUI automation, several limitations warrant discussion. First, our current implementation of exploration anchors relies on mobile app manifest declarations (e.g., Android Activity components), which limits direct applicability to web and desktop environments. Second, although the current Knowledge Ranker takes only 28.5 seconds per step, it's still a bit slow. Future work will focus on extending this approach to web and desktop and speeding up Knowledge Ranker.

%每次生成任务的时候并不能总是保证mllm能生成max\_branching\_factor个任务，有的页面可能就没有任务，例如进入了about页面。不过点应该不算是我们的Limitations
%不是所有的状态都能成功恢复，例如某个任务是在YouTube指定的视频下发送指定的评论，那么它的兄弟任务的初始环境状态就无法恢复到未发送评论的状态。这个例子也不太好用，毕竟是可以通过点击删除按钮来撤销的，但是实际上我们难以（因为要考虑到整个state space）用一个预先写好的状态恢复方法去消除/重现所有动态生成的任务对环境的影响。
%\C{To be continued. See https://aclrollingreview.org/cfp\#limitations }
%\C{Authors are required to discuss the limitations of their work in a dedicated section titled “Limitations”. This section should be included at the end of the paper, before the references, and it will not count toward the page limit. This includes both, long and short papers. Papers without a limitations section will be desk rejected.}
%\C{Please note that this section should not introduce new methods, analysis, or results. We reserve the right to desk reject the submissions that use this section to introduce more content that should have been part of the main paper. It can only discuss the limitations of the work presented in the main content of the paper.}

\section*{Ethics Statement}
%\C{To be continued. See https://aclrollingreview.org/cfp\#ethics-policy and https://aclrollingreview.org/responsibleNLPresearch }
%可以参考 https://arxiv.org/pdf/2310.05007 这样声明吗？

Our work introduces GUI-explorer, an autonomous agent for graphical user interface automation, and raises several ethical considerations inherent to AI-driven interaction systems. First, while our exploration process utilizes application screenshots and accessibility metadata, we strictly employ open-source or publicly available applications, ensuring no collection of private user data or infringement of intellectual property rights. 

Second, our reliance on large multimodal models introduces potential risks of perpetuating societal biases embedded in their training data. Though our transition-aware knowledge mechanism mitigates the misinterpretation of UI components, we acknowledge that residual biases in element interpretation could lead to unintended operational consequences. We strongly advocate for human oversight in real-world deployments, particularly for sensitive applications in healthcare or finance domains.

The computational costs associated with our approach (average 66 seconds per interaction step) raise environmental concerns regarding energy consumption. While our method eliminates the need for model retraining—a significant carbon footprint contributor—future work must prioritize efficiency optimizations to enable sustainable scaling.

We recognize potential dual-use risks where autonomous GUI agents could be misused for malicious automation (e.g., credential stuffing or click fraud), much like other AI technologies can be used for creating deceptive presentations or face presentation attacks~\citep{shao2019multi,shao2025deepfake}. 

Finally, our benchmark construction followed ethical annotation practices, with contributors compensated at fair market rates and granted full rights to withdraw their participation. 

%\section*{Acknowledgements}

%We thank anonymous reviewers for their constructive feedback. This study is supported by National Natural Science Foundation of China (Grant No. 62306090), Natural Science Foundation of Guangdong Province of China (Grant No. 2024A1515010147), China Postdoctoral Science Foundation (Grant No. 2024M764192) and Shenzhen Science and Technology Program (KQTD20240729102207002).

\bibliography{custom}

%\newpage
\clearpage
\appendix
\section{Hardware Configurations}
\label{sec:Hardware-configurations}
Hardware configurations were optimized for cost-effectiveness: Most experiments ran on a single NVIDIA GeForce RTX 4070 Laptop GPU (8GB VRAM). For GUI-KRB evaluations involving open-source MLLMs, we scaled to two NVIDIA L40S GPUs (48GB VRAM) to accommodate larger VRAM requirements.

\section{GUI-KRB Benchmark Distributions}
\label{sec:GUI-KRB-Distributions}
\begin{figure}[htb]
  \includegraphics[width=\columnwidth]{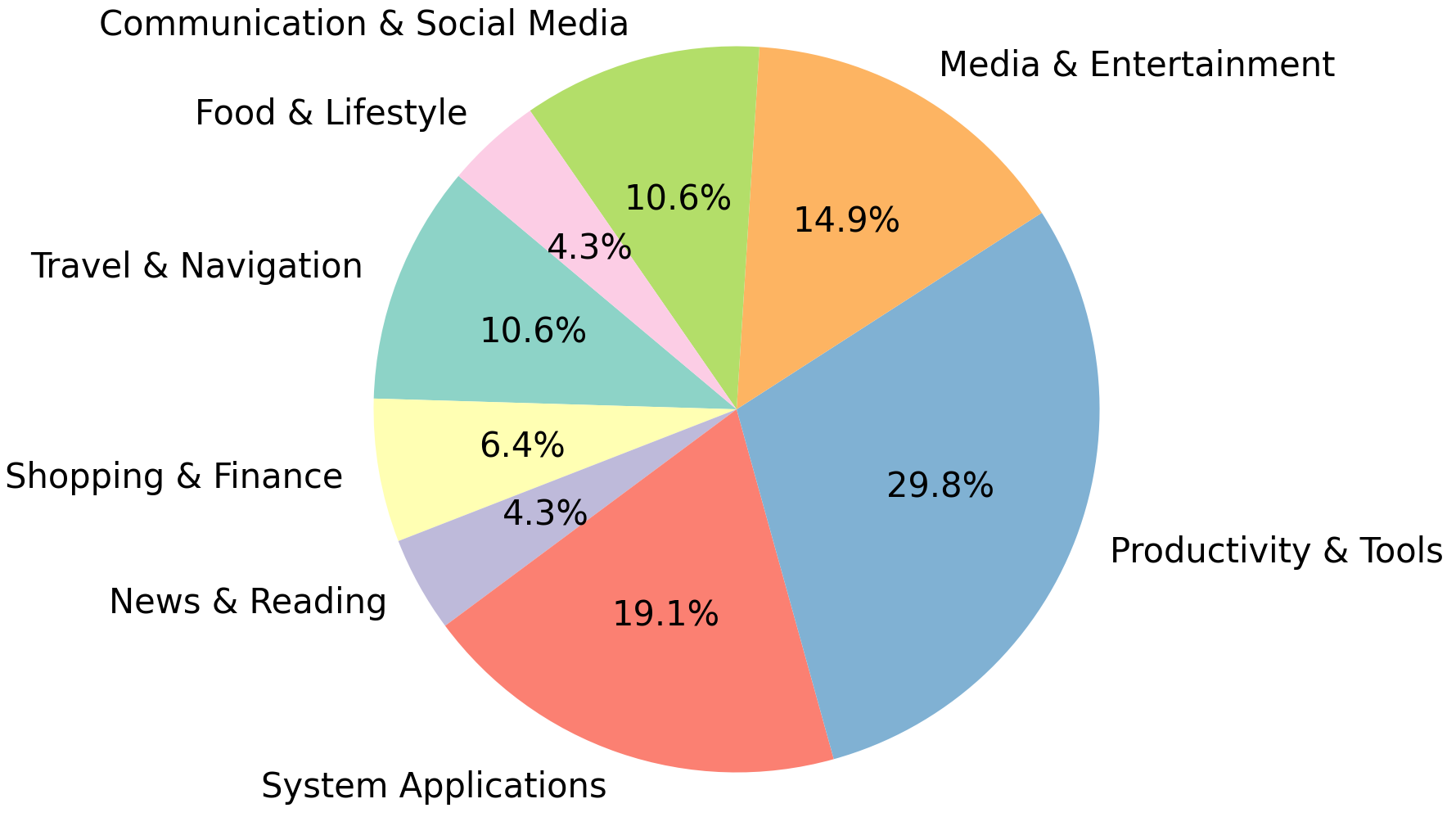}
  \caption{Distribution of apps in GUI-KRB.}
  \label{fig:app_category_distribution}
  %\vspace*{-3mm}
\end{figure}

Figure~\ref{fig:app_category_distribution} shows the distribution of the number of apps in GUI-KRB.

\section{GUI-KRB Benchmark Sample Data}
\label{sec:GUI-KRB-sample}
\begin{figure}[htb]
  \includegraphics[width=\linewidth]{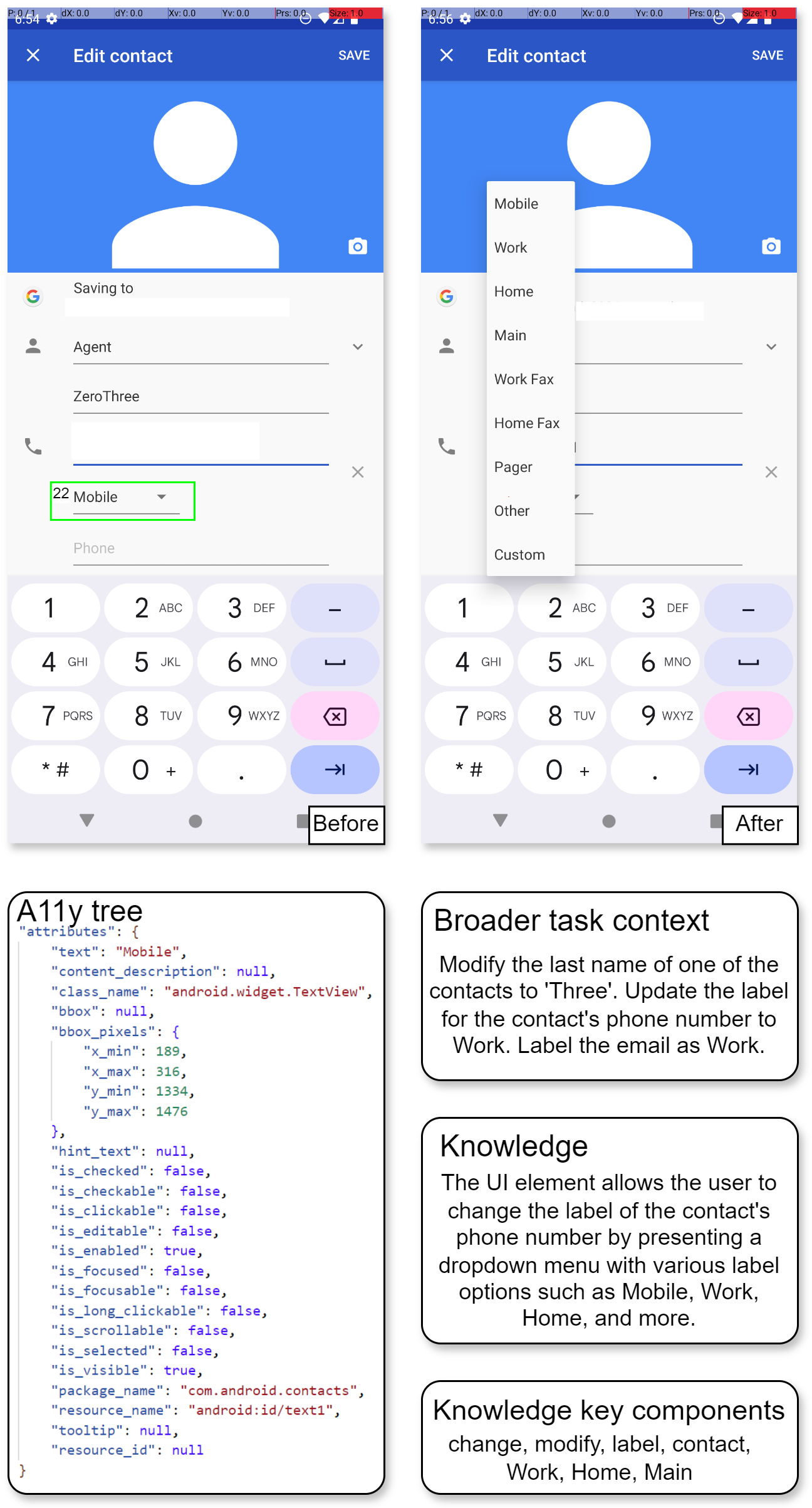}
  \caption{A comprehensive sample from GUI-KRB benchmark illustrating: (1) before/after screenshots of target element interaction, (2) accessibility tree representation, (3) broader task context, (4) transition-aware knowledge (excluded from test input), and (5) evaluation keywords with synonyms for robust assessment.}
  \label{fig:GUI-KRB-sample}
  %\vspace*{-3mm}
\end{figure}

This section presents an example from the GUI-KRB benchmark to illustrate its data structure, as shown in Figure~\ref{fig:GUI-KRB-sample}. Each sample consists of five main components. First, it contains screenshots captured before and after the interaction with the target element to demonstrate the visual state transition. Second, it includes the accessibility tree representation of the interface. Third, the broader task context describes the necessary interaction with the target element required to complete the task. Fourth, the transition-aware knowledge associated with the element is documented but excluded from the test input. Finally, for automated evaluation purposes, the sample includes evaluation keywords (also excluded from test input) that incorporate synonyms and related terms (such as "modify" and "Main") to accommodate various valid responses and reduce false judgments during model assessment.

\section{Understanding Exploration Anchors}
\label{sec:Understanding-Exploration-Anchors}
As introduced in \secref{sec:Autonomous-Exploration-of-Function-aware-Trajectory}, Exploration Anchors are key entry points to application functionalities. These are explicitly declared by developers within the app’s manifest file (e.g., Android’s \texttt{AndroidManifest.xml}) and typically correspond to \textit{activities}, which represent individual screens or distinct functional units within the app.

\subsection{Extraction and Utilization}
Exploration Anchors are systematically extracted from the app's manifest file, which enumerates all developer-declared activities. During the autonomous exploration process, the agent utilizes these anchors to formulate intermediate navigational goals. For example, if an anchor corresponding to a specific activity (e.g., an activity named \texttt{ShareInstagramStory}) is not directly accessible from the agent's current screen, the agent will iteratively generate sub-goals, such as navigating through menus or interacting with UI elements, to eventually reach the target anchor activity.

\subsection{Illustrative Case Studies}
To further clarify the concept and utility of Exploration Anchors, consider the following examples from the ``Retro Music'' application. Its manifest file declares activities including \texttt{ShareInstagramStory}, \texttt{DriveModeActivity}, and \texttt{RestoreActivity}. If the agent begins its exploration from a screen displaying ``Most played'' tracks, it might generate intermediate goals such as:
\begin{enumerate}
    \item ``Tap `Most played', then share the top song to Instagram Story.'' \\
          (Targeting anchor: \texttt{ShareInstagramStory})
    \item ``Tap the settings icon, then navigate to Drive Mode and enable it.'' \\
          (Targeting anchor: \texttt{DriveModeActivity})
\end{enumerate}

\section{Error Analysis}
In this section, we categorize and discuss three primary error types observed in our evaluation trajectories, detailing their component-level manifestations and root causes.

\subsection{Perceptual Errors}
Perceptual errors occur when agents misinterpret visual or contextual cues. For instance, in the SPA-Bench \texttt{dictionary\_merriam\_webster\_2} task, the agent failed to recognize that a solid icon indicated an already ``saved'' word. Instead, it redundantly clicked the ``save'' button, unintentionally unsaving the word. Such errors often stem from limitations in grounding GUI elements (e.g., distinguishing icon states like filled versus hollow) or parsing dynamic UI hierarchies (e.g., overlays, scrolling content).

\subsection{Reasoning Errors}
Reasoning errors arise from incomplete task decomposition or flawed step-by-step logic. For example, in SPA-Bench \texttt{contacts\_2}, the agent appended ``Three'' to an existing last name instead of first deleting the original text, thereby violating the task’s implicit requirement to ``replace'' rather than ``modify''. These errors reflect limitations in the base model’s ability to infer nuanced constraints (e.g., distinguishing ``edit'' from ``overwrite'') or manage multi-step dependencies (e.g., ensuring changes are saved before exiting).

\subsection{Missing Knowledge Errors}
Missing knowledge errors occur when agents lack app-specific prior knowledge critical for task completion. For instance, in SPA-Bench \texttt{booking.com\_5}, the agent exhausted its step limit searching for a ``currency'' setting under ``Preferences'' instead of the correct ``Payment details'' section. This highlights challenges in efficiently navigating unfamiliar UIs, particularly when key functionalities are nested within non-intuitive menus.

\section{Comparison with Alternative Ranking Methods}
To validate the necessity and efficiency of our pairwise knowledge ranking module, we conducted comparative experiments against three alternative ranking strategies:
(1) Direct LLM judgment: The LLM directly assesses the usefulness of each knowledge piece.
(2) Confidence scores: Ranking based on confidence scores assigned to knowledge.
(3) Sliding-window ranking~\citep{sun-etal-2023-chatgpt}: A sequential ranking approach.

We evaluated these methods on the GUI-KRB benchmark using an identical experimental setup (Qwen2-VL-72B-Instruct-GPTQ-Int4). 
\begin{table}[htbp]%\small
\centering
%\footnotesize
\resizebox{\columnwidth}{!}{
\begin{tabular}{lcc}
\toprule[1.5pt]
\textbf{Ranking Method} & \textbf{\begin{tabular}[c]{@{}c@{}}Prior Knowledge \\ Error Rate (\%)\end{tabular}} & \textbf{\begin{tabular}[c]{@{}c@{}}Dynamic Comprehen- \\ sion Error Rate (\%)\end{tabular}}\\
\midrule
Direct LLM judgment & 8.8 & 8.8 \\
Confidence score & 7.6 & 7.0 \\
Sliding-window~\citep{sun-etal-2023-chatgpt} & 6.8 & 6.4 \\
%\midrule
\rowcolor[HTML]{E7EEFE}
\textbf{Ours} & \textbf{6.8} & \textbf{6.4} \\
\bottomrule[1.5pt]
\end{tabular}
}
\caption{Comparison of Different Knowledge Ranking Methods on GUI-KRB. Error rates (\%) are reported. Lower is better.}
\label{tab:ranking_comparison}
%\vspace*{-6mm}
\end{table}
The results in Table~\ref{tab:ranking_comparison} demonstrate that both direct LLM judgment and confidence-based scoring yield higher error rates than our proposed method. While sliding-window ranking achieves comparable error rates to our approach, it suffers from a time complexity of $O(n)$ due to its sequential traversal of knowledge items. In contrast, our merge-sort-inspired pairwise comparison strategy allows for $O(\log n)$ time complexity through parallelizable divide-and-conquer iterations. This inherent parallelism makes our method significantly more scalable for large knowledge sets. For instance, to reduce latency with sliding-window ranking, one would need to compromise sorting quality (e.g., by enlarging window or step sizes). Crucially, the computational overhead of merge operations in our method is negligible, as LLM-based pairwise comparisons (which incur seconds-level delays) dominate the overall runtime.

Furthermore, as shown in Table~\ref{tab:results-gui-krb}, disabling ranking increases the Prior Knowledge error rate from 6.8\% to 9.8\% , highlighting the critical role of ranking in performance.

\section{Generalization and Robustness Analysis}
\label{sec:generalization_robustness}
This section presents additional experimental results to address the robustness of our autonomous exploration approach with incomplete metadata and the generalization of our framework to web environments.

\subsection{Robustness to Incomplete Metadata}
\label{sec:robustness_incomplete_metadata}
To assess the performance of our autonomous exploration approach when application metadata is incomplete or unavailable, we conducted supplementary experiments where exploration task goal generation relied solely on screenshot data, without access to manifest-derived structural information (Exploration Anchors as described in \secref{sec:Understanding-Exploration-Anchors}). This scenario simulates conditions where only visual information is accessible.

We tested this screenshot-only configuration on the Retro Music and Broccoli applications, evaluating performance using the corresponding task sets from the AndroidWorld benchmark. The results, compared against a baseline without any exploration and our full method (which utilizes manifest metadata), are presented in Table~\ref{tab:robustness_metadata}.

\begin{table}[htb]
\centering
\resizebox{\columnwidth}{!}{%
\begin{tabular}{lc}
\toprule[1.5pt]
\textbf{Method Configuration} & \textbf{Task Success Rate (\%)} \\
\midrule
No exploration (Baseline)      & 16.7                 \\
Ours (Screenshots only) & 22.2                 \\
Ours (Full method)      & 33.3                 \\
\bottomrule[1.5pt]
\end{tabular}%
}
\caption{Performance comparison on selected AndroidWorld tasks (Retro Music and Broccoli apps) under varying levels of metadata availability for exploration task goal generation.}
\label{tab:robustness_metadata}
\end{table}

The results indicate that even when restricted to using only screenshots for generating exploration goals, our method achieves a 5.5\% absolute improvement in task success rate over the non-exploratory baseline. While the integration of full metadata (Exploration Anchors) clearly yields superior performance (33.3\% success rate), these experiments confirm that the core exploration strategy remains effective and provides benefits even in metadata-deficient scenarios. This demonstrates a degree of robustness in our approach, as the visual cues present in screenshots can still guide meaningful exploration, albeit less efficiently than when supplemented with structural priors from manifest files. 

\subsection{Generalization to Web Environments}
\label{sec:generalization_web}
While our method, particularly the Autonomous Exploration component described in \secref{sec:Autonomous-Exploration-of-Function-aware-Trajectory}, leverages Android-specific structural information (i.e., Exploration Anchors from manifest files) to guide exploration, it is pertinent to investigate the potential for the acquired knowledge to generalize to other platforms, such as web applications.

To this end, we conducted additional experiments to evaluate the cross-platform applicability of the transition-aware knowledge mined from Android environments. Specifically, we applied our framework, using knowledge acquired solely from Android app exploration, to tasks in a web environment. The evaluation was performed on the "website" split of the Multimodal-Mind2Web test set, comprising 20 distinct samples. We compared the performance of a baseline agent without any augmented knowledge against an agent augmented with static knowledge derived from our Android exploration phase. The results are presented in Table~\ref{tab:generalization_web}.

\begin{table}[htb]
\centering
\resizebox{\columnwidth}{!}{%
\begin{tabular}{lcc}
\toprule[1.5pt]
\textbf{Knowledge Configuration} & \textbf{\begin{tabular}[c]{@{}c@{}}Macro Element \\ Accuracy (\%)\end{tabular}} & \textbf{\begin{tabular}[c]{@{}c@{}}Macro Step \\ Accuracy (\%)\end{tabular}} \\
\midrule
No augmented knowledge (Baseline) & 45.97                    & 42.31                 \\
+ Static Android-derived knowledge & \textbf{47.26}                & \textbf{43.05}             \\
\bottomrule[1.5pt]
\end{tabular}%
}
\caption{Performance on the Multimodal-Mind2Web "website" test split (20 samples) with and without leveraging static knowledge acquired from Android app exploration.}
\label{tab:generalization_web}
\end{table}

The results show a modest improvement in both macro element accuracy (+1.29\%) and macro step accuracy (+0.74\%) when leveraging knowledge acquired from Android applications. This suggests that some fundamental aspects of UI interaction logic, such as understanding hierarchical structures, common action sequences (e.g., "select then confirm"), and visual-textual correlations, possess a degree of cross-platform generalizability. While these improvements are not as substantial as those observed within the Android domain, they indicate that the core principles of transition-aware knowledge are not strictly confined to mobile environments and can offer some benefit in web contexts without any web-specific exploration or fine-tuning.

It is important to note that web environments present unique challenges, such as highly dynamic content, complex DOM structures, and browser-specific interactions, which are not fully addressed by knowledge solely derived from Android apps. Future work will focus on adapting the autonomous exploration and knowledge mining mechanisms specifically for web and desktop platforms to achieve more significant performance gains. However, these initial findings provide encouraging evidence of the underlying generalizability of the learned GUI interaction patterns.

\section{Distribution of Transition-aware Knowledge}
\label{sec:Distribution-of-Transition-aware-Knowledge}
\begin{figure}[htb]
  %\vspace*{-3mm}
  \includegraphics[width=\linewidth]{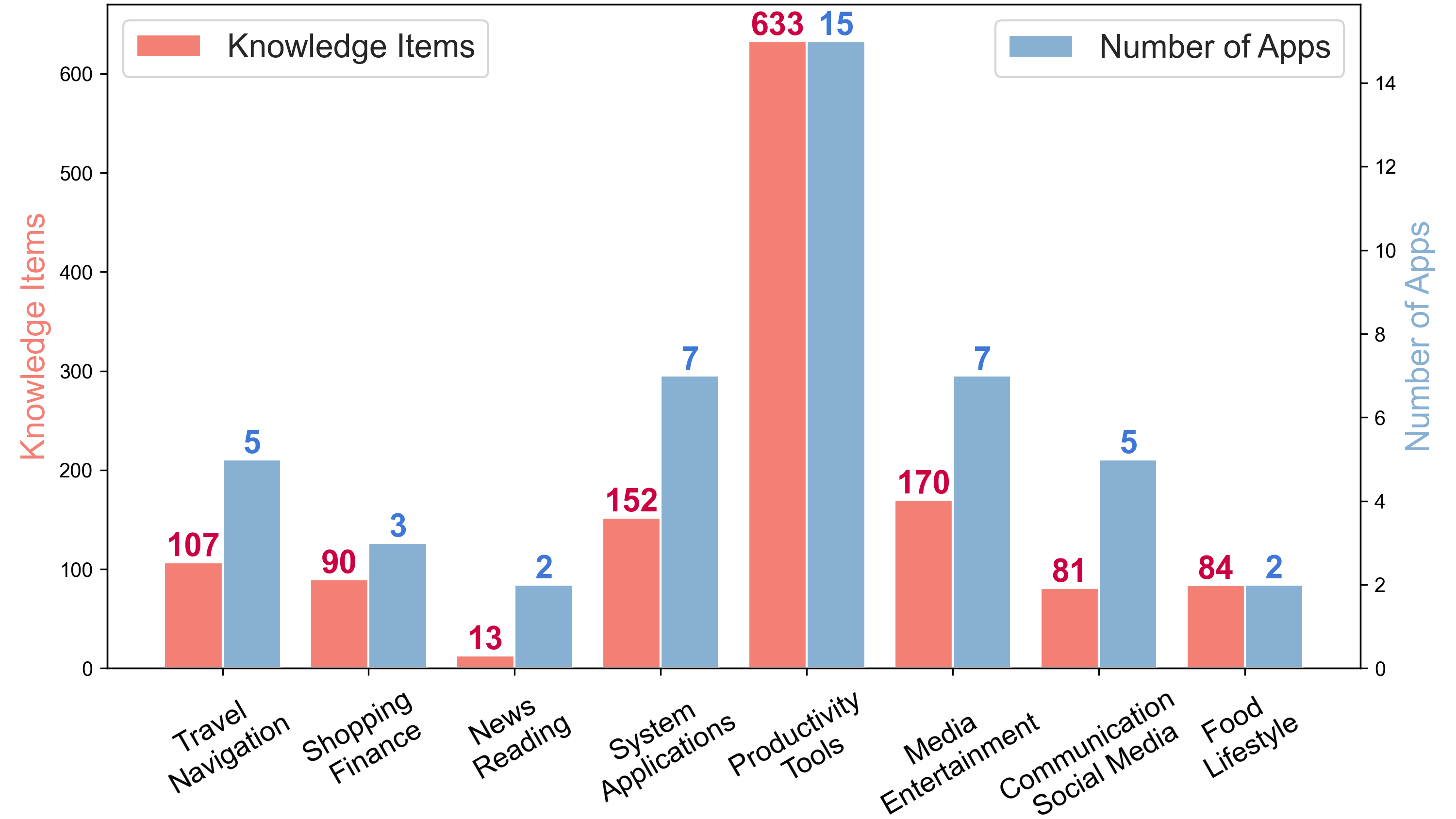}
  \caption{Distribution of transition-aware knowledge gained through autonomous exploration.}
  \label{fig:distribution_of_knowledge_and_number_of_apps}
  %\vspace*{-3mm}
\end{figure}
Figure~\ref{fig:distribution_of_knowledge_and_number_of_apps} shows the distribution of transition-aware knowledge gained through autonomous exploration.

\section{Prompting Templates of GUI-explorer}
\subsection{Prompting Template of Function-aware Task Goal Generator}
\label{prompt:task-goal-generator}
%\begin{tcolorbox}[mypromptstyle]
\begin{prompt}
Given the screenshot of \arginputs{{app name}} and its available activities, generate a comprehensive list of practical user tasks that:

1. Start from the current screen shown in the screenshot
2. Can be completed within 10-30 steps
3. Utilize the app's full feature set based on the activity list
4. Are concrete and specific (like searching for a particular item rather than just "search")
5. Cover different user interaction patterns (viewing, editing, sharing, etc.)
6. Include both basic and advanced features
7. Represent realistic user behaviors and goals
8. Avoid excessive steps on form-filling or scrolling pages
\end{prompt}
\begin{prompt}
Important context:
- App name: \arginputs{{app name}}
- Package name: \arginputs{{package name}}
- Available activities (app screens/features):
\arginputs{{activity list}}

Format requirements:
1. List only the tasks without explanations or commentary
2. Each task should be a single, clear directive
3. Use specific examples (e.g., concrete search terms, actions, settings)
4. Include the expected outcome where relevant
5. Tasks should follow this pattern: [Starting action] + [Specific steps] + [End goal]

Example tasks from other apps (for reference only):
1. Search for "ocean waves" white noise, then sort results by most played
2. Open the first recommended video, then post "Great content!" as a comment
3. Play the trending video, then add it to your "Watch Later" playlist
4. Navigate to the comments section of a featured video, then like the top comment

Generate diverse tasks that would help a user explore and utilize all major features visible in the screenshot and implied by the activity list.
\end{prompt}
%\end{tcolorbox}

\subsection{Prompting Template of Unsupervised Mining of Transition-aware Knowledge}
\label{prompt:transition-aware-knowledge}
\begin{prompt}
Objective: Describe the functionality of a specific UI element in a mobile app screenshot.

Input:
- Two screenshots: Before and after interacting with a UI element
- UI element marked with a numeric tag in the top-left corner
- Element number: \arginputs{{numeric tag of element}}
- Broader task context: \arginputs{{task description}}
- Action taken: \arginputs{{action}}
- UI Element Attributes: 
  ```
  \arginputs{{ui element attributes}}
  ```

Requirements for Functionality Description:
1. Concise: 1-2 sentences
2. Focus on general function, not specific details
3. Avoid mentioning the numeric tag
4. Use generic terms like "UI element" or appropriate pronouns

Example:
- Incorrect: "Tapping the element \#3 displays David's saved recipes in the results panel"
- Correct: "Tapping this element will initiates a search and displays matching results"

Guidance:
- Describe the core action and immediate result of interacting with the UI element
- Prioritize clarity and generality in the description
\end{prompt}

\subsection{Prompting Template of Knowledge Ranker}
\label{prompt:knowledge-ranker}
\begin{prompt}
Given the user instruction: \arginputs{{task goal}}, determine which of the following two knowledge entries is more useful.
Respond ONLY with a integer value:
1 means Knowledge A is strictly better.
2 means Knowledge B is strictly better.

Knowledge A: \arginputs{{knowledge a}}
Knowledge B: \arginputs{{knowledge b}}

Please provide your response:
\end{prompt}

\subsection{Prompting Template of Reasoning}
\begin{prompt}
\#\# Role Definition
You are an Android operation AI that fulfills user requests through precise screen interactions.
The current screenshot and the same screenshot with bounding boxes and labels added are also given to you.

\#\# Action Catalog
Available actions (STRICT JSON FORMAT REQUIRED):
1. Status Operations:
   - Task Complete: \{"action\_type": "status", "goal\_status": "complete"\}
   - Task Infeasible: \{"action\_type": "status", "goal\_status": "infeasible"\}
2. Information Actions:
   - Answer Question: \{"action\_type": "answer", "text": "<answer\_text>"\}
3. Screen Interactions:
   - Tap Element: \{"action\_type": "click", "index": <visible\_index>\}
   - Long Press: \{"action\_type": "long\_press", "index": <visible\_index>\}
   - Scroll: Scroll the screen or a specific scrollable UI element. Use the `index` of the target element if scrolling a specific element, or omit `index` to scroll the whole screen. {{"action\_type": "scroll", "direction": <"up"|"down"|"left"|"right">, "index": <optional\_target\_index>}}
4. Input Operations:
   - Text Entry: \{"action\_type": "input\_text", "text": "<content>", "index": <text\_field\_index>\}
   - Keyboard Enter: \{"action\_type": "keyboard\_enter"\}
5. Navigation:
   - Home Screen: \{"action\_type": "navigate\_home"\}
   - Back Navigation: \{"action\_type": "navigate\_back"\}
6. System Actions:
   - Launch App: \{"action\_type": "open\_app", "app\_name": "<exact\_name>"\}
   - Wait Refresh: \{"action\_type": "wait"\}

\#\# Current Objective
User Goal: \arginputs{{task goal}}

\#\# Execution Context
Action History:
\arginputs{{history}}

Visible UI Elements (Only interact with *visible=true elements):
\arginputs{{ui elements}}

\#\# Core Strategy
1. Path Optimization:
   - Prefer direct methods (e.g., open\_app > app drawer navigation)
   - Always use the `input\_text` action for entering text into designated text fields.
   - Verify element visibility (`visible=true`) before attempting any interaction (click, long\_press, input\_text). Do not interact with elements marked `visible=false`.
   - Use `scroll` when necessary to bring off-screen elements into view. Prioritize scrolling specific containers (`index` provided) over full-screen scrolls if possible.

2. Error Handling Protocol:
   - Switch approach after $\geq 2$ failed attempts
   - Prioritize scrolling (`scroll` action) over force-acting on invisible elements
   - If an element is not visible, use `scroll` in the likely direction (e.g., 'down' to find elements below the current view).
   - Try opposite scroll direction if initial fails (up/down, left/right)
    - If the `open\_app` action fails to correctly open the app, find the corresponding app in the app drawer and open it.

3. Information Tasks:
   - MANDATORY: Use answer action for questions
   - Verify data freshness (e.g., check calendar date)

\#\# Expert Techniques
Here are some tips for you:
\arginputs{{knowledge}}

\#\# Response Format
STRICTLY follow:
Reasoning: [Step-by-step analysis covering:
           - Visibility verification
           - History effectiveness evaluation
           - Alternative approach comparison
\end{prompt}
\begin{prompt}
           - Consideration of scrolling if needed]
Action: [SINGLE JSON action from catalog]

Generate response:
\end{prompt}

\section{Prompting Templates of GUI-KRB}
\subsection{Prompting Template of Prior Knowledge Task}
\begin{prompt}
Objective: Describe the functionality of a specific UI element in a mobile app screenshot.

Input:
- One screenshot: Before interacting with a UI element
- UI element marked with a numeric tag in the top-left corner
- Element number: \arginputs{{numeric tag of element}}
- Broader task context: \arginputs{{task description}}
- UI Element Attributes: 
  ```
  \arginputs{{ui element attributes}}
  ```
  
Requirements for Functionality Description:
1. Concise: 1-2 sentences
2. Focus on general function, not specific details
3. Avoid mentioning the numeric tag
4. Use generic terms like "UI element" or appropriate pronouns

Example:
- Incorrect: "Tapping the element \#3 displays David's saved recipes in the results panel"
- Correct: "Tapping this element will initiates a search and displays matching results"

Guidance:
- Describe the core action and immediate result of interacting with the UI element
- Infer functionality based on the current screen context
- Prioritize clarity and generality in the description
\end{prompt}

\subsection{Prompting Template of Dynamic Comprehension Task}
Same as Appendix~\ref{prompt:transition-aware-knowledge}.

\subsection{Prompting Templates for GUI-explorer (w/o Ranker)}
\subsubsection{Prompting Template of Prior Knowledge Task}
\begin{prompt}
Objective: Describe the functionality of a specific UI element in a mobile app screenshot.

Input:
- One screenshot: Before interacting with a UI element
- UI element marked with a numeric tag in the top-left corner
- Element number: \arginputs{{numeric tag of element}}
- Broader task context: \arginputs{{task description}}
- UI Element Attributes: 
  ```
  \arginputs{{ui element attributes}}
  ```
- Similar UI Elements' Functionalities (retrieved based on visual similarity):
  ```
  \arginputs{{similar element functionalities}}
  ```

Requirements for Functionality Description:
1. Concise: 1-2 sentences
2. Focus on general function, not specific details
3. Avoid mentioning the numeric tag
\end{prompt}
\begin{prompt}
4. Use generic terms like "UI element" or appropriate pronouns
5. Consider similar elements' functionalities as reference, but prioritize:
   - Current screen context
   - UI element attributes
   - Task description
6. Only incorporate relevant patterns from similar elements if they align with the current context

Example:
- Incorrect: "Tapping the element \#3 displays David's saved recipes in the results panel"
- Correct: "Tapping this element will initiates a search and displays matching results"

Guidance:
- Describe the core action and potential result of interacting with the UI element
- Infer functionality based on the current screen context
- Prioritize clarity and generality in the description
- Use similar elements' functionalities to validate and refine your description, not to simply copy them
\end{prompt}

\subsubsection{Prompting Template of Dynamic Comprehension Task}
\begin{prompt}
Objective: Describe the functionality of a specific UI element in a mobile app screenshot.

Input:
- Two screenshots: Before and after interacting with a UI element
- UI element marked with a numeric tag in the top-left corner
- Element number: \arginputs{{numeric tag of element}}
- Broader task context: \arginputs{{task description}}
- UI Element Attributes: 
  ```
  \arginputs{{ui element attributes}}
  ```
- Similar UI Elements' Functionalities (retrieved based on visual similarity):
  ```
  \arginputs{{similar element functionalities}}
  ```

Requirements for Functionality Description:
1. Concise: 1-2 sentences
2. Focus on general function, not specific details
3. Avoid mentioning the numeric tag
4. Use generic terms like "UI element" or appropriate pronouns
5. Consider similar elements' functionalities as reference, but prioritize:
   - Current screen context
   - UI element attributes
   - Task description
6. Only incorporate relevant patterns from similar elements if they align with the current context

Example:
- Incorrect: "Tapping the element \#3 displays David's saved recipes in the results panel"
- Correct: "Tapping this element will initiates a search and displays matching results"

Guidance:
- Describe the core action and immediate result of interacting with the UI element
- Infer functionality based on the current screen context
- Prioritize clarity and generality in the description
- Use similar elements' functionalities to validate and refine your description, not to simply copy them
\end{prompt}

\subsection{Prompting Templates of Prior Knowledge Task for GUI-explorer}
\subsubsection{Prompting Template of Prior Knowledge Task}
\begin{prompt}
Objective: Describe the functionality of a specific UI element in a mobile app screenshot.

Input:
- One screenshot: Before interacting with a UI element
- UI element marked with a numeric tag in the top-left corner
- Element number: \arginputs{{numeric tag of element}}
- Broader task context: \arginputs{{task description}}
- UI Element Attributes: 
  ```
  \arginputs{{ui element attributes}}
  ```
- Similar UI Elements' Functionalities (ranked by relevance to task description):
  ```
  \arginputs{{similar element functionalities}}
  ```
  Note: Elements are sorted by relevance, with most task-relevant functionalities listed first

Requirements for Functionality Description:
1. Concise: 1-2 sentences
2. Focus on general function, not specific details
3. Avoid mentioning the numeric tag
4. Use generic terms like "UI element" or appropriate pronouns
5. Consider similar elements' functionalities as reference, with priority:
   - Higher-ranked (more relevant) reference functionalities
   - Current screen context
   - UI element attributes
   - Task description
6. Only incorporate relevant patterns from similar elements if they align with the current context

Example:
- Incorrect: "Tapping the element \#3 displays David's saved recipes in the results panel"
- Correct: "Tapping this element will initiates a search and displays matching results"
Guidance:
- Describe the core action and potential result of interacting with the UI element
- Infer functionality based on the current screen context
- Prioritize clarity and generality in the description
- Pay special attention to higher-ranked similar functionalities as they are more likely to be relevant
- Use similar elements' functionalities to validate and refine your description, not to simply copy them
\end{prompt}

\subsubsection{Prompting Template of Dynamic Comprehension Task}
\begin{prompt}
Objective: Describe the functionality of a specific UI element in a mobile app screenshot.

Input:
- Two screenshots: Before and after interacting with a UI element
- UI element marked with a numeric tag in the top-left corner
- Element number: \arginputs{{numeric tag of element}}
- Broader task context: \arginputs{{task description}}
- UI Element Attributes: 
  ```
  \arginputs{{ui element attributes}}
  ```
- Similar UI Elements' Functionalities (ranked by relevance to task description):
  ```
  \arginputs{{similar element functionalities}}
  ```
  Note: Elements are sorted by relevance, with most task-relevant functionalities listed first
\end{prompt}
\begin{prompt}
Requirements for Functionality Description:
1. Concise: 1-2 sentences
2. Focus on general function, not specific details
3. Avoid mentioning the numeric tag
4. Use generic terms like "UI element" or appropriate pronouns
5. Consider similar elements' functionalities as reference, with priority:
   - Higher-ranked (more relevant) reference functionalities
   - Current screen context
   - UI element attributes
   - Task description
6. Only incorporate relevant patterns from similar elements if they align with the current context

Example:
- Incorrect: "Tapping the element \#3 displays David's saved recipes in the results panel"
- Correct: "Tapping this element will initiates a search and displays matching results"

Guidance:
- Describe the core action and potential result of interacting with the UI element
- Infer functionality based on the current screen context
- Prioritize clarity and generality in the description
- Pay special attention to higher-ranked similar functionalities as they are more likely to be relevant
- Use similar elements' functionalities to validate and refine your description, not to simply copy them
\end{prompt}

\end{document}